\documentclass{article}

\usepackage{arxiv}

\usepackage{amsmath} 

\usepackage[utf8]{inputenc} 
\usepackage[T1]{fontenc}    
\usepackage{hyperref}       
\usepackage{url}            
\usepackage{booktabs}       
\usepackage{amsfonts}       
\usepackage{nicefrac}       
\usepackage{microtype}      
\usepackage{cleveref}       
\usepackage{graphicx}
\usepackage[numbers]{natbib}
\usepackage{doi}

\usepackage{multirow}%
\usepackage{amssymb}%
\usepackage{amsthm}%
\usepackage{mathrsfs}%
\usepackage[title]{appendix}%
\usepackage{xcolor}%
\usepackage{textcomp}%
\usepackage{manyfoot}%
\usepackage{booktabs}%
\usepackage{algpseudocode}%
\usepackage{listings}%
\usepackage{makecell}%
\usepackage[table]{xcolor}%
\usepackage{longtable}

\title{Extracting ontology-compliant knowledge from scientific text describing irradiated materials using large language models}

\date{}

\newif\ifuniqueAffiliation
\uniqueAffiliationtrue

\ifuniqueAffiliation 
\author{
    Marco Luca Sbodio\thanks{These authors contributed equally to this work.}\\
    IBM Research\\
    \texttt{marco.sbodio@ie.ibm.com}\\
    \And
    Marcos Martínez Galindo\footnotemark[1]\\
    IBM Research\\
    \texttt{Marcos.Martinez.Galindo@ibm.com}\\
    \And
    Vanessa Lopez\footnotemark[1]\\
    IBM Research\\
    \texttt{vanlopez@ie.ibm.com}\\
    \And
    Blanca Biel\\
    Dpto. Física Atómica, Molecular y Nuclear\\
    Instituto Carlos I de Física Teórica y Computacional\\
    Univ. of Granada, Spain\\
    \texttt{biel@ugr.es}\\
    \And
    Pablo Canca\\
    Dpto. Física Atómica, Molecular y Nuclear\\
    Univ. of Granada, Spain\\
    \texttt{pcanca@ugr.es}\\
    \And
    Pedro Delgado\\
    IFMIF DONES, Spain\\
    \texttt{pedro.delgado@ifmif-dones.es}\\
    \And
    Jesús I. Mendieta-Moreno\\
    Instituto de Ciencia de Materiales de Madrid (ICMM)\\
    CSIC, Spain\\
    \texttt{jmendi85@gmail.com}\\
    \And
    Raphael Tack\\
    IBM Research\\
    \texttt{raphael.tack@ibm.com}\\
    \And
    Maria J. Caturla\\
    Dpto. de Física, Facultad de Ciencias\\
    Universidad de Alicante, Spain\\
    \texttt{mj.caturla@gcloud.ua.es}\\
}
\else
\usepackage{authblk}

\newbox{\orcid}\sbox{\orcid}{\includegraphics[scale=0.06]{orcid.pdf}} 
\fi

\renewcommand{\undertitle}{}
\renewcommand{\headeright}{}
\renewcommand{\shorttitle}{Extracting ontology-compliant knowledge from scientific text describing irradiated materials using LLMs}

\hypersetup{
pdftitle={Extracting ontology-compliant knowledge from scientific text describing irradiated materials using large language models},
pdfsubject={cs.AI},
pdfauthor={Marco Luca Sbodio, Marcos Martínez Galindo, Vanessa Lopez, Blanca Biel, Pablo Canca, Pedro Delgado, Jesús I. Mendieta-Moreno, Raphael Tack, Maria J. Caturla},
pdfkeywords={Knowledge Graph, Large Language Models, LLM, information extraction, irradiation defects, fusion materials},
}

\begin{document}
\maketitle

\begin{abstract}
The quest for new materials increasingly relies on predictive models and comprehensive simulations that span scales from atomic to macroscopic levels. However, essential data necessary for these models and simulations are often embedded in scientific literature as unstructured text, limiting reusability and posing challenges for researchers seeking to leverage existing knowledge effectively. While extracting structured data from unstructured text using large language models is gaining popularity, traditional methods typically generate key-value pairs data with straightforward schemas. In contrast, we introduce \textit{eolas}, a modular pipeline that uses large language models to automatically transform scientific documents into knowledge graphs aligned with a specified ontology. We demonstrate \textit{eolas} effectiveness in extracting useful information for scientists studying materials designed to endure the extreme temperatures and radiation levels found in fusion reactors. While a human expert might spend between thirty to ninety minutes extracting relevant data from an article, \textit{eolas} can generate high-quality knowledge graphs in just a few minutes. These are presented in a tabular format with faceted navigation for easy human validation. Additionally, we introduce the first benchmark dataset designed to assess large language models’ capabilities in constructing knowledge graphs within the domain of irradiated materials. The analysis of 168 experiments using our dataset, various large language models and prompting techniques provides key insights that we summarize into practical guidelines for effectively extracting knowledge graphs aligned with an input ontology.
\end{abstract}

\section{Introduction}

The demand for novel materials to address critical scientific challenges is rapidly increasing. In particular, nuclear fusion, a global research priority with its potential to provide nearly unlimited clean energy, requires identifying materials with high radiation and temperature resistance \cite{WAS2019_fission_fusion, PINTSUK2022_review}. This extreme environment, coupled with the technical difficulties of experimental validation, make exhaustive physical verification of every prospective material infeasible. Initiatives like IFMIF-DONES (International Fusion Materials Irradiation Facility—Demo-Oriented Neutron Source) \cite{Ibarra_2018}, currently under construction, aim to advance experimental capabilities through high-energy neutron irradiation. Besides such experimental facilities, research in the domain relies on computer-assisted discovery, through simulations and verification processes leveraging prior scientific knowledge.

\begin{figure}
\centering
\includegraphics[width=\textwidth]{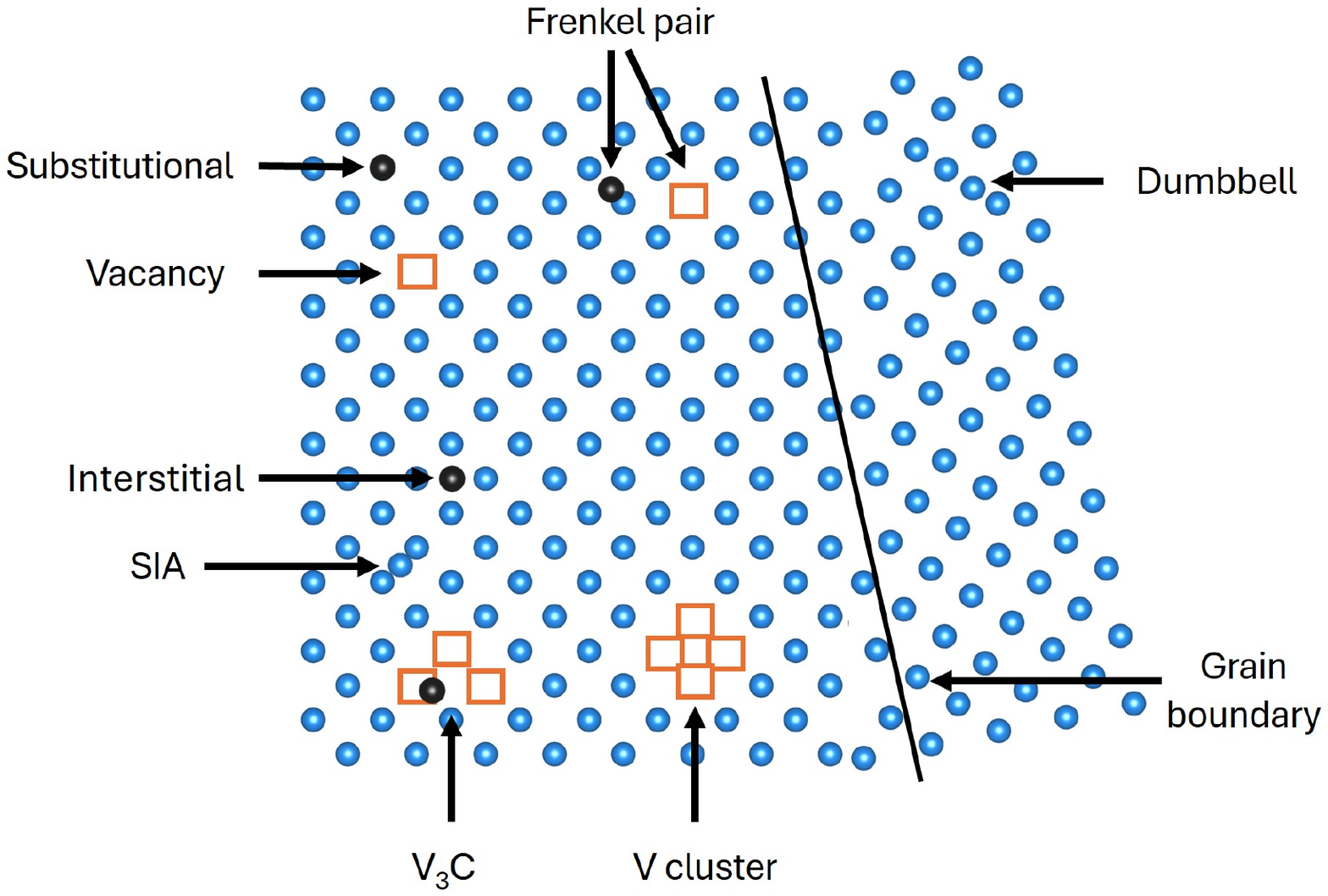}
\caption{Illustration of some of the most representative defects observed in materials under irradiation conditions: vacancies, where an atom is missing from a lattice site; self-interstitial (SIA) or foreign interstitials, where an atom occupies a site between the perfect crystal lattice sites; Frenkel pair, a vacancy and its corresponding interstitial atom; dumbbells, where two atoms share the same lattice site; or bigger defect structures, such as clusters or grain boundaries. \\ 
}
\label{defects_example}
\end{figure}

Arguably, one of the main challenges in modeling materials' responses to irradiation is accounting for the multitude of complex variables involved. These include the composition of the material, the types of defects produced, and the specific irradiation conditions, all of which are crucial for accurately predicting long-term macroscopic effects. Defects are formed at the atomic level and can be quite complex. Figure~\ref{defects_example} illustrates a few examples of atomic-scale defects, ranging from simple types such as vacancies (missing atoms in a crystal lattice) and interstitials (atoms positioned outside perfect lattice sites), to more complex structures like clusters of point defects and grain boundaries. Each exhibits different geometries that influence the mechanical, electronic, and magnetic properties of materials. Some properties of interest of these defects are their formation energy, which indicates relative stability; migration energy, which reveals their ability to diffuse within the lattice driven by temperature; or their relaxation volume, which quantifies the elastic deformation they induce. Researchers have extensively studied these properties using atomic-scale models at varying levels of accuracy, ranging from Density Functional Theory (DFT) \cite{hohenberg1964density, kohn1965self} to classical molecular dynamics (MD) \cite{MALERBA2021_parameters}. The findings are widely documented across numerous sources. Parameters from atomic scale models are subsequently integrated into long-term microstructural evolution simulations \cite{DUDAREV2025_review, MALERBA2021_M4F, LASA2024_review} using a multiscale modeling approach. However, retrieving, comparing, and classifying the information already available in the literature presents significant challenges.

Scientific discovery fundamentally relies on prior domain knowledge, particularly in AI-assisted research \cite{Wang2023ScientificDI}. Large language models (LLMs) have significantly advanced text-based knowledge extraction \cite{survey_llms_kg_construction, Text2KGBench, Dagdelen2024, Polak_2024, dasilva2024automatedllmenabledextraction}. However, the use of generative AI for evaluating materials designed for future fusion reactors remains largely unexplored. A key obstacle is the limited availability of structured data, as crucial information is often embedded in text or tables. This makes it difficult to consolidate knowledge and develop predictive models.

Motivated by this high-impact application, we demonstrate how LLMs, when combined with a domain ontology, can effectively extract structured, machine-readable information about atomistic defect modeling in irradiated materials from the literature, with minimal human intervention.

LLMs, pre-trained on extensive unlabeled datasets through self-supervised learning, can be adapted for specific tasks via fine-tuning or in-context learning (ICL). In ICL \cite{NEURIPS2020_1457c0d6}, a prompt comprising instructions and examples guides the model without modifying its parameters. Typically, methods using pre-trained LLMs employ pipelines that perform named entity and relation extraction from general-purpose text at sentence or paragraph level. These approaches may utilize prompts containing examples (known as few-shot) to guide the model in extracting entities and relations \cite{itext2kg}. Alternatively, they might leverage open-domain ontologies to direct LLMs in extracting simple triples ($<$subject, predicate, object$>$) \cite{Text2KGBench, khorashadizadeh2023} as opposed to fully connected knowledge graphs.

In materials science, MaterioMiner \cite{Durmaz2024AnOT} conducts named-entity recognition to annotate literature with a domain-specific ontology focused on material fatigue. 
Acknowledging the complexity inherent in extracting expert knowledge from scientific literature, there is a growing recognition of the need for more flexible, schema-driven approaches. Recent efforts extend the extraction to increasingly complex structures. For instance, the approach described in \cite{Dagdelen2024} fine-tunes LLMs to produce structured lists of user-defined JSON objects, incorporating a predefined set of keys tailored for materials chemistry tasks. In another approach, \texttt{ChatExtract} \cite{Polak_2024} utilizes a conversational LLM with zero-shot prompting to extract material property triplets in the format of $<Material, Value, Unit>$; the proposed method enhances accuracy through a series of follow-up questions. Additionally, KEP \cite{dasilva2024automatedllmenabledextraction} introduces a pipeline for extracting synthesis protocols specific to reticular materials; it uses few-shot prompts and automatically selects the most suitable examples to guide an LLM in generating outputs encoded in JSON format. Collectively, these methodologies, whether employing fine-tuning or prompt-based ICL, underscore the significant potential of LLMs in automating domain-specific knowledge extraction guided by schemas, going beyond mere entity and pariwise relation extraction.

In this study, we developed an ontology to define complex semantic relationships within the domain of defect energetics in irradiated materials. We derived its structure and content from a thorough review of relevant domain literature and insights gained through interviews with scientists working in the field. Leveraging RDF \cite{schreiber2014primer} and OWL \cite{OWL} standards, the ontology establishes a common vocabulary that supports semantic interoperability and facilitates the transformation of raw text into meaningful structured knowledge. 

\begin{figure}
    \centering
    \includegraphics[height=0.75\textheight, keepaspectratio]{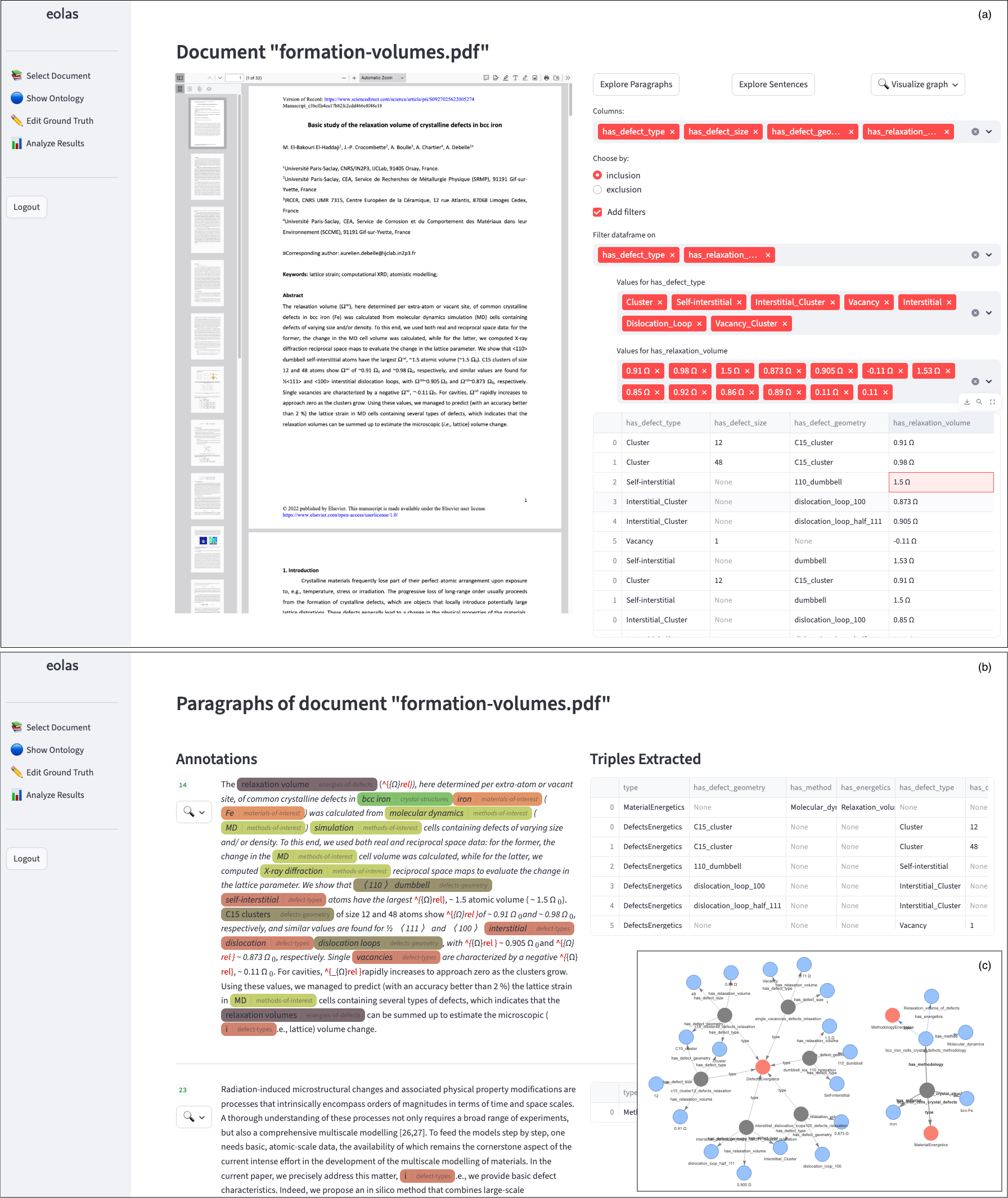}
    \caption{Screenshots of the \textit{eolas} User Interface. (a) The interface displays the knowledge graph extracted from a document \cite{relaxation} (left) as tabular data (right). This table can be queried and filtered according to user needs. (b) Users can explore extraction results at a finer level by viewing tabular data corresponding to specific text fragments, with an optional visualization of the underlying knowledge graph (c). This semantically-rich knowledge graph captures measurements from multiple defect types, sizes and geometries for a material of interest, extracted from a text passage containing numerous interrelated entities and relations. The visualization shows relations between intermediate entities (grey nodes), the ontology classes they instantiate (pink nodes), and their associated values or leaf entities (blue nodes). These relationships describe: (1) materials and their crystal structure (eg., bcc Iron) (2) measurements (e.g., relaxation volume) for the different defect types and geometries for a given material;  and (3) the methodology and parameters used to compute these values (e.g, Molecular Dynamics). The extracted knowledge for that passage is presented into a table ((b) top right), where leaf properties appear as columns and the connected intermediate nodes are represented within the same row. This detailed view allows for more precise analysis and understanding of individual components within the extracted data.}
    \label{fig:diagrams-eolas_user_interface_with_graph}
\end{figure}

We present \textit{eolas}, our domain-agnostic modular pipeline designed to automate extraction of knowledge from text in accordance with an input ontology. Named after the Irish word for "knowledge," \textit{eolas} consolidates data into a semantically compliant knowledge graph (KG) at document level. This KG can be queried, serialised (e.g, in turtle syntax) or displayed as structured tabular data in \textit{eolas} user interface, accompanied by the original text passages from which it was extracted. This feature facilitates inspection and validation by domain experts, enhancing transparency and accuracy (see figure \ref{fig:diagrams-eolas_user_interface_with_graph}). A demo version of \textit{eolas} user interface is available at \url{https://demo-public.1fotembijozx.eu-es.codeengine.appdomain.cloud/}.

Based on our ontology, we present the first benchmark dataset to evaluate the capabilities of LLMs in extracting ontology-compliant KGs in this complex domain. The versatility and modularity of the \textit{eolas} pipeline enabled us to efficiently conduct 168 experiments based on our benchmark. Using the \textit{eolas} pipeline, we experimented with various openly available LLMs, different (automatically constructed) ontology-guided prompting techniques, diverse representations of our ontology schema, and heuristics for semantically post-processing model outputs. For few-shot prompting, a small ontology-complaint KG extracted from a handful of passages serves as the reference input–output examples —each pairing a text passage with its corresponding KG — used to guide the model at inference time (ICL). Since these KG examples must be manually created and reviewed (in its tabular form) by domain experts, their quantity is limited, making them insufficient for fine-tuning a model. 
We present the results of our experiments, describe our findings on LLMs' capabilities for ontology-guided knowledge extraction, and summarize lessons learned and opportunities.

The ontology and benchmark datasets are open source (\url{https://github.com/jmendi/eolas_irradiated_materials/}) to encourage further research. Though demonstrated here within the context of irradiated materials, both the ontology and benchmark capture essential information about atomic-scale defects. This makes them broadly applicable to various other material domains. Similarly, our  \textit{eolas} pipeline for extracting knowledge graphs from text can be adapted across diverse fields since all domain-specific details are encapsulated in the ontology.

\section{Results}

\subsection{Benchmark dataset}

We provide the first benchmark dataset to evaluate the capabilities of LLMs in extracting knowledge graphs in the domain of materials for future fusion reactors. We built our benchmark data through a semi-automated iterative approach, involving both subject matter experts (SMEs) and LLMs. The following are the main steps for the construction of the benchmark data.
\begin{itemize}
    \item We manually defined an ontology schema to model the semantics of concepts and relationships needed to describe defect energetics in irradiated materials, as reported in scientific literature. Optionally, some ontology classes are associated with non-exhaustive dictionaries of known entity instances (e.g, materials, defect types, defect geometries or methods).
    \item SMEs selected five representative scientific articles in the target domain \cite{multiscale-modelling-irradiated, relaxation, PhysRevB.92.104102, barouh_PhysRevB.90.054112, MALERBA2021_parameters}, and manually identified 111 relevant text passages (a mix of sentences, paragraphs, and one table with caption). We added 15 randomly selected irrelevant passages, resulting in a set $\mathcal{G}_{initial}$ (initial ground truth) consisting of 126 text passages.
    \item We manually chose a set $\mathcal{E} \subset \mathcal{G}_{initial}$ containing 15 example passages: 14 are relevant, and 1 is irrelevant. Using the Protégé ontology editing environment \cite{protege}, we manually created a small KG compliant with our ontology for these 14 examples. These passages were carefully chosen for their diversity and representativeness, ensuring that collectively, their corresponding KGs encompass most of the classes and relationships defined in the ontology.
    \item We defined an initial prompt configuration to instruct an LLM to extract a KG in the form of turtle triples \cite{w3Turtle} from a given passage of text. The prompt configuration includes an instruction, a representation of the ontology, the set $\mathcal{E}$ of few shot examples, each associated with a serialization in turtle format of its corresponding KG (the irrelevant passage is associated with "NA" to instruct the LLM to return this string rather than generating a KG for irrelevant passages). The prompt is used to extract a KG for each of the remaining 111 target passages of text in $ \mathcal{G}_{initial} \setminus \mathcal{E}$.
    \item We built a web application that converts the extracted KGs into tables (each row represents an entity, and the columns are the properties), and displays them along with the passage of text. A team of five domain experts and 3 researchers used our web interface to curate the tables (change/add/delete values and/or rows), and finally validate them.
    \item Finally, we converted the validated tables back into KGs, and stored them along with the corresponding passages of text. This semi-automated approach, in which the ground truth is collaboratively generated by LLMs and SMEs, proved to be more efficient than manually populating the tables from scratch.
\end{itemize}

The process was iterative. Early iterations revealed that domain experts found the initial ontology schema insufficient for representing all necessary domain knowledge without ambiguity. Consequently, we refined the ontology and adjusted both the few-shot examples in the knowledge graph and the ground truth using the web application. This refinement enabled us to precisely and unambiguously capture the types and relationships consistent with the ontology across 126 passages that constitute our ground truth dataset.

The final benchmark dataset $\mathcal{G}$ comprises 126 text passages, each paired with a KG. However, for evaluation purposes, the 15 passages used in few-shot prompting are excluded. This results in an evaluation set comprising 111 passages.

The final version of the ontology schema consists of 14 classes, 111 instances, 13 object properties, and 25 data properties. The KG with the 14 examples in $\mathcal{E}$ yields 299 turtle triples with 75 unique entities and 36 unique properties, and all the 126 KGs in $\mathcal{G}$ yield 1314 turtle triples, with 317 unique entities and 32 unique properties. We publicly release both the ontology, and the benchmark dataset at \url{https://github.com/jmendi/eolas_irradiated_materials/}.

\subsection{Description of the experiments}

The experiments utilize our benchmark dataset and our \textit{eolas} pipeline to assess the performance of different LLMs and different prompting techniques. More precisely, we use 6 openly accessible LLMs: OpenAI gpt-oss-120b and gpt-oss-20b \cite{openai2025gptoss120bgptoss20bmodel}, IBM granite-3.3-8b-instruct \cite{huggingfaceIbmgranitegranite338binstructHugging}, Meta llama-3-3-70b-instruct \cite{huggingfaceMetallamaLlama3370BInstructHugging} and llama-4-maverick-17b-128e-instruct-fp8 \cite{huggingfaceMetallamaLlama4Maverick17B128EInstructFP8Hugging}, Mistral AI mistral-large \cite{huggingfaceMistralaiMistralLargeInstruct2407Hugging}; for simplicity, we will refer to these models as $\text{GPT\_120, GPT\_20, GRANITE, LLAMA\_3, LLAMA\_4, and MISTRAL}$, respectively. For all LLMs, we use the same set of common parameters, including a fixed random seed and $temperature = 0.0$ to reduce randomness. In our experiments, we use configurable prompts designed to guide the model in extracting data from provided texts based on a specified ontology schema. The configuration is designed to be agnostic of both the specific domain and ontology, and it allows \textit{eolas} to dynamically generate prompts as needed.

We explore various prompt configurations by varying their complexity and format. Specifically, the instructions within these prompts can be either simple or detailed. Detailed instructions offer more precise guidance to the model by directing it to generate turtle triples exclusively based on the context passage and ontology, to ensure accuracy and completeness, adherence to domain/range and functional constraints, disregard information that is not present in the text, and use consistent and unique URIs. 
Additionally, the prompt contains the ontology schema, which can be serialized into three distinct formats: turtle triples syntax (TTL) \cite{w3Turtle}, BAML \cite{Boundary_BAML_-_A_2025} (a language with a compact syntax for defining types), and a simple textual verbalization (VERB), listing the classes and the relations (including their domains) using just their labels. Furthermore, our approach includes an option to incorporate instances from the ontology: these predefined individuals are associated (through dictionaries) with classes within the ontology schema, enriching the domain's vocabulary and helping the LLM to map textual mentions to known instances with unambiguous URIs. We also examine the impact of including examples in the prompts. This involves comparing few-shot learning setups, where examples are provided, against zero-shot scenarios, which do not include any examples. Finally, we evaluate the impact of some post-processing heuristics to consolidate the output of the LLM. 

The few-shot prompts use the examples in the set $\mathcal{E} \subset \mathcal{G}$, which contains 14 relevant examples (passage of text associated with a manually curated KG) and 1 irrelevant example (a passage of text associated with the string 'NA'). To ensure a fair comparison between few-shot and zero-shot experiments, we conduct all experiments exclusively on the dataset $\mathcal{G} \setminus \mathcal{E} = \left[ \langle t_i, G^E_i \rangle, i \in \left[1, 111\right] \right]$, where $\mathcal{G}$ is our benchmark dataset, $t_i$ is a text passage, and $G^E_i$ is the expected KG associated with $t_i$; $\mathcal{G} \setminus \mathcal{E}$ is an array with 111 items.

By combining the different prompt configurations with the various LLMs, we have 168 experimental configurations (see table \ref{table:experiments_summary}). Given a configuration $c_j$, running an experiment yields the results $\mathcal{R}_j = \left[ G^P_{i,j} = f \left( c_j, t_i \right) \mid \langle t_i, G^E_i \rangle \in \mathcal{G} \setminus \mathcal{E} \right]$, where $G^P_{i, j}$ is a predicted KG extracted ($f$) from the text $t_i$ using the instructions, serialization, instances, examples, and consolidation specified by configuration $c_j$. $\mathcal{R}_j$ is an array with $n = 111$ items.

\begin{table}
    \centering
    \begin{tabular}{cccccc}
         \toprule
         \makecell[c]{instructions,\\ serialization} & use instances & use examples & use consolidation & LLMs & \makecell[c]{number of\\ configurations} \\
         \midrule
         S, BAML & $\left\{T, F\right\}$  & $\left\{T, F\right\}$ & $\left\{T, F\right\}$ & $\mathcal{L}$ & 48 \\
         S, TTL & $\left\{T, F\right\}$ & $\left\{T, F\right\}$ & $\left\{T, F\right\}$ & $\mathcal{L}$ & 48 \\
         D, TTL & $\left\{T, F\right\}$  & $\left\{T, F\right\}$ & $\left\{T, F\right\}$ & $\mathcal{L}$ & 48 \\
         S, VERB & $\left\{F\right\}$  & $\left\{T\right\}$ & $\left\{T, F\right\}$ & $\mathcal{L}$ & 12 \\
         D, VERB & $\left\{F\right\}$  & $\left\{T\right\}$ & $\left\{T, F\right\}$ & $\mathcal{L}$ & 12 \\
         \bottomrule
    \end{tabular}
    \caption{Summary of the 168 experiments using 6 different LLMs ($\mathcal{L} = \left\{ \text{GPT\_120, GPT\_20, GRANITE, LLAMA\_3, LLAMA\_4, MISTRAL} \right\}$), and different prompt structures. The prompt may include simple (S) or detailed (D) instructions, and a serialization of the ontology schema in either BAML, TTL or a verbalized (VERB) textual format. The prompt may optionally (True, False) include ontology instances, and examples (few-shot vs zero-shot prompts). Finally, we may (True, False) use some heuristics to consolidate LLM results.}
    \label{table:experiments_summary}
\end{table}

\subsection{Evaluation metrics}

\begin{figure}
    \centering
    \includegraphics[width=\linewidth]{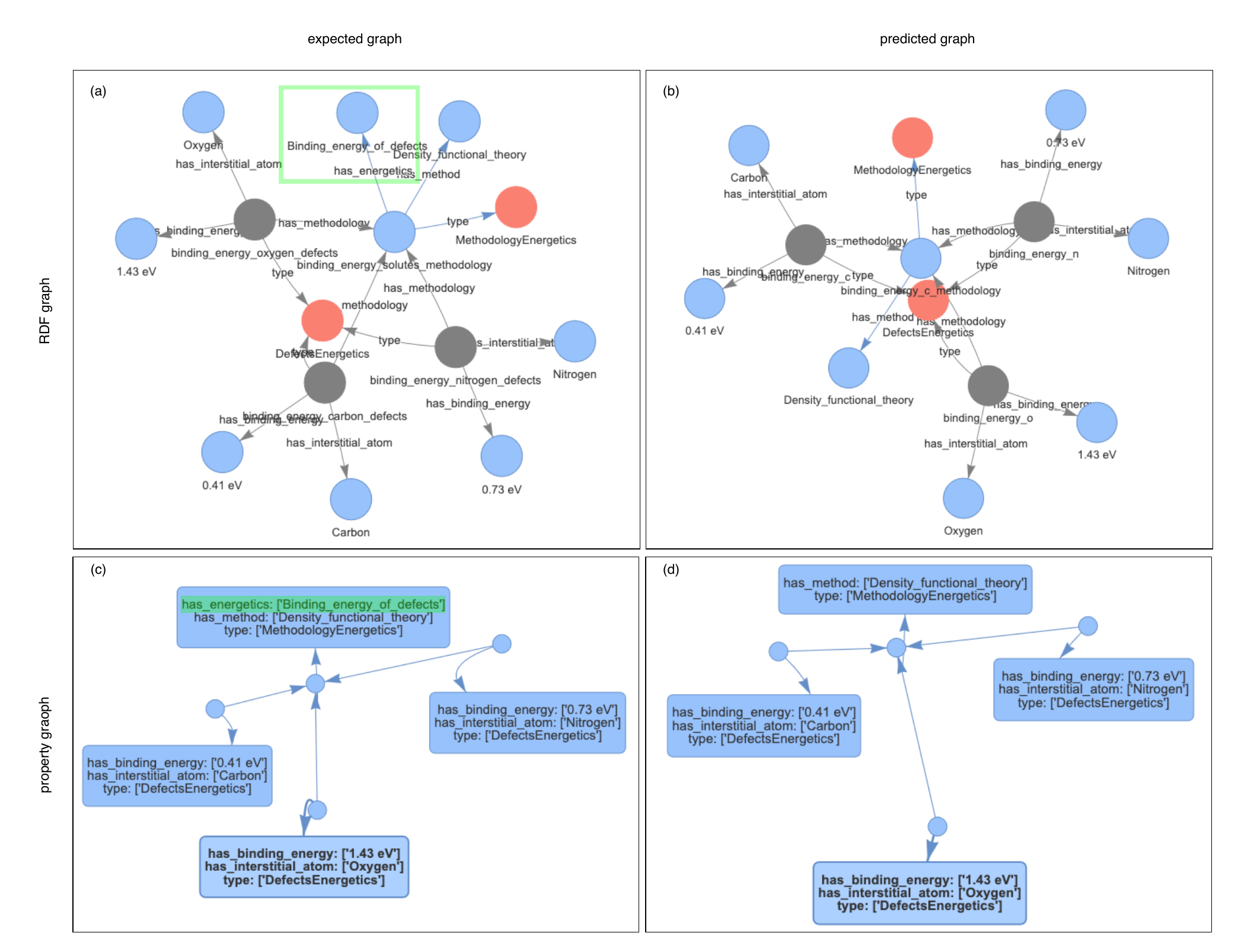}
    \caption{To mitigate the computational cost of the graph edit distance, we transform KGs from RDF format (a, b) into property graphs (c, d), which have fewer nodes/edges. The example shows the expected graph (a, c) corresponding to the following fragment of text "The corresponding binding energies are 1.43, 0.73, and 0.41 eV for the case of O, N, and C, respectively, in good agreement with previous DFT values [6,17,18,22]" (from \cite{barouh_PhysRevB.90.054112}). The predicted graph (b, d) is generated using LLAMA\_3 with a few-shot prompt, detailed instructions, ontology schema serialized in turtle format (TTL), with no instances, and using consolidation. We highlight in green in (a, c) the node and edge (property and value) that are missing in the predicted graph, resulting in a graph edit distance of 2.}
    \label{fig:diagraph-GED}
\end{figure}

To evaluate the results of an experimental configuration $c_j$, we need to compare each predicted KG $G^P_{i, j}$ with the corresponding expected KG $G^E_i$ from the benchmark dataset. The Graph Edit Distance ($GED$)~\cite{ged} is a metric widely used to compare graphs: it measures the minimum cost of converting one graph into the other by using a set of graph edit operations, typically addition (of a node or an edge), deletion (of a node or an edge), and  substitution (of a node or edge with another node or edge, respectively). Each graph edit operation has a cost. Computation of the exact $GED$ is an NP-Hard problem~\cite{Zeng2009}, which makes it impractical for large graphs. KGs, typically represented using RDF \cite{schreiber2014primer}, have a large number of nodes and edges because all properties of an entity (node), including literal values (such as numerical or string values), are also represented as nodes (for a formal definition of RDF literals see \cite{RDFConceptsAbstract}). To mitigate this computational problem, we transform RDF graphs (both expected and predicted) into property graphs \cite{angles2018property} (see figure \ref{fig:diagraph-GED}), where each node may have properties (or attributes). We use the properties of a node in the property graph to represent the literal nodes in the RDF graph, and we use a different cost for the edit operations (add/delete/substitute), taking into account the number of properties in the node being added/deleted/substituted. This transformation reduces the number of nodes and edges, enabling the computation of $GED$ within a reasonable time frame (for the actual computation of $GED$ we use the \texttt{graph\_edit\_distance} function from the NetworkX library \cite{SciPyProceedings_11}).

The graph edit distance is expressed as a natural number. However, comparing these values can be challenging because context matters. For instance, consider two expected graphs $ G^E_a $ and $ G^E_b
$, with 2 and 20 nodes respectively, alongside their corresponding predicted graphs $ G^P_{a,j} $ and $ G^P_{b,j} $. We might find that $ GED(G^E_a, G^P_{a,j}) = 1 < GED(G^E_b, G^P_{b,j})= 2 $, yet, intuitively, a graph edit distance of 1 on a small graph with only 2 nodes is worse than a graph edit distance of 2 on a larger graph with 20 nodes. This suggests that the LLM using configuration $ c_j $ performed better predicting $ G^P_{b,j} $ compared to $ G^P_{a,j} $.
To address these issues, we adopt the Normalized Graph Edit Distance (\(NGED\)), which is a rational number ranging from 0 to 1; a value of 0 indicates that the two graphs being compared are identical (best scenario). To normalize the graph edit distance, we make the following assumption: given an expected graph \( G^E_i \) and its corresponding predicted graph \( G^P_{i, j} \), the most costly transformation involves two steps. First, convert \( G^P_{i, j} \) into the empty graph \( G^{\emptyset} \) (a graph with no nodes or edges), and then transform \( G^{\emptyset} \) into \( G^E_i \). Consequently, the maximum cost to transform \( G^P_{i, j} \) into \( G^E_i \) is given by \( \text{GED}(G^P_{i, j}, G^{\emptyset}) + \text{GED}(G^{\emptyset}, G^E_i) \). The normalized graph edit distance is then the fraction of this maximum cost that we actually incur when transforming \( G^P_{i, j} \) into \( G^E_i \).

\begin{equation}
\label{eq:NGED}
    NGED(G^P_{i, j}, G^E_i) = 
    \frac{
        GED(G^P_{i, j}, G^E_i)
    }
    {
        GED(G^P_{i, j}, G^{\emptyset}) + 
        GED(G^{\emptyset}, G^E_i)
    }
\end{equation}

Analyzing two extreme cases where \(NGED(G^P_{i, j}, G^E_i) = 1\) provides valuable insights. In the first scenario, if \(G^E_i = G^{\emptyset}\), then \(GED(G^{\emptyset}, G^E_i) = 0\); consequently, \(GED(G^P_{i, j}, G^E_i) = GED(G^P_{i, j}, G^{\emptyset})\). Here, the LLM with configuration \(c_j\) predicts a non-empty graph, while the expected result is an empty graph. This represents a scenario where the LLM's prediction consists entirely of hallucinations \cite{kalai2025languagemodelshallucinate}.
In the second scenario, if \(G^P_{i, j} = G^{\emptyset}\), then \(GED(G^P_{i, j}, G^{\emptyset}) = 0\). Therefore, \(GED(G^P_{i, j}, G^E_i) = GED(G^{\emptyset}, G^E_i)\). In this case, the LLM predicts an empty graph while a non-empty graph was expected. This illustrates a situation where the LLM fails to extract any relevant data from the provided text.

For each experimental configuration $c_j$, we compute its results $\mathcal{R}_j$; for every predicted graph $G^P_{i, j} \in \mathcal{R}_j$, we calculate its normalized graph edit distance ($NGED$) from the corresponding expected graph $G^E_i \in \mathcal{G} \setminus \mathcal{E}$. This procedure produces the array $NGED_j$, with values ranging between 0 and 1:

\begin{equation}
\label{eq:NGED_j}
    NGED_j = \left[ NGED(G^P_{i, j}, G^E_i) \mid G^P_{i, j} \in \mathcal{R}_j \wedge G^E_i \in \mathcal{G} \setminus \mathcal{E} \right]    
\end{equation}

To assign a score to a configuration $c_j$, we set a threshold value $NGED^*$ for the normalized graph edit distance, and we compute the percentage of values in $NGED_j$ that are less than or equal to $NGED^*$. We refer to this score as $P_j(NGED^*)$, the percentage of success at $NGED^*$:

\begin{equation}
\label{eq:percentage-of-success}    
    P_j(NGED^*) = 
    \frac{
        | \left\{x \le NGED^* \mid x \in NGED_j \right\} |
    }
    {
        | NGED_j |
    }
\end{equation}

Intuitively, \(P_j(NGED^*)\) represents the percentage of test cases within the benchmark dataset \(\mathcal{G} \setminus \mathcal{E}\) for which configuration \(c_j\) predicts a graph that we can transform into the expected graph with an incurred cost that does not exceed \(NGED^*\) times the maximum possible cost. Given a value $NGED^*$, a configuration \(c_j\) is considered better than another configuration \(c_k\) if \(P_j(NGED^*) > P_k(NGED^*)\). The selection of the threshold value \( NGED^* \) indicates how much error is tolerated during evaluation: choosing lower values for \( NGED^* \) results in stricter evaluations and reduces tolerance for mistakes made by LLMs when extracting knowledge graphs.

We use the SciPy library \cite{2020SciPy-NMeth} and specifically the \texttt{percentileofscore} function (with parameter \texttt{kind='strict'}) to compute \(P_j(NGED^*)\).

\subsection{Results of the experiments}

Tables \ref{table:ranking-based-on-percentage-of-success-at-nged-0.1} and \ref{table:ranking-based-on-percentage-of-success-at-nged-0.2} report the percentage of success for 168 \textit{eolas }configurations evaluated on the 111 passages in the ground truth set, using thresholds $NGED^* = 0.1$ and $NGED^* = 0.2$, respectively (appendix \ref{sec:appendix:Percentage of success at different thresholds} reports results with other values of $NGED^*$). The verbalized (VERB) schema, which simply lists the names of relevant classes and their relations, lacks sufficiently detailed descriptions, making it effective only when accompanied by examples in prompts. Additionally, this type of serialization does not accommodate inclusion of instances. Consequently, we report results for VERB serialization exclusively under few-shot configurations that do not incorporate instances explicitly (aside from those provided in the few-shot examples themselves). 


\begin{table}
\centering
\caption{Percentage of success 
at normalized graph edit distance $NGED^* = 0.1$ 
for all experimental configurations. Each cell reports the score $P_j(0.1)$ 
(see equation \ref{eq:percentage-of-success}) for the configuration as a percentage.
Colors indicate \colorbox[HTML]{ FFF4B8 }{best}, 
\colorbox[HTML]{ DBDBDB }{2nd best}, and \colorbox[HTML]{ F1DBC6 }{3rd best} 
for each column.}
\label{table:ranking-based-on-percentage-of-success-at-nged-0.1}
\begin{tabular}{rr|rrrr|rrrr}
\toprule
 & & \multicolumn{4}{c|}{zero-shot} & \multicolumn{4}{c}{few-shot} \\
\midrule
 & examples & \multicolumn{4}{c|}{False} & \multicolumn{4}{c}{True} \\
 & instances & \multicolumn{2}{c}{False} & \multicolumn{2}{c|}{True} & \multicolumn{2}{c}{False} & \multicolumn{2}{c}{True} \\
 & consolidation & False & True & False & True & False & True & False & True \\
\makecell[r]{instructions\\serialization} & LLM &  &  &  &  &  &  &  &  \\
\midrule
\multirow[c]{6}{*}{S, TTL} & GPT\_120 & {\cellcolor[HTML]{FFF4B8}} 12.6 & {\cellcolor[HTML]{FFF4B8}} 12.6 & {\cellcolor[HTML]{DBDBDB}} 12.6 & 12.6 & {\cellcolor[HTML]{DBDBDB}} 37.8 & {\cellcolor[HTML]{F1DBC6}} 38.7 & 30.6 & 32.4 \\
 & GPT\_20 & {\cellcolor[HTML]{FFF4B8}} 12.6 & {\cellcolor[HTML]{FFF4B8}} 12.6 & {\cellcolor[HTML]{DBDBDB}} 12.6 & 12.6 & 9.0 & 9.0 & 7.2 & 8.1 \\
 & GRANITE & 4.5 & 5.4 & 1.8 & 2.7 & 13.5 & 13.5 & 10.8 & 11.7 \\
 & LLAMA\_3 & {\cellcolor[HTML]{DBDBDB}} 11.7 & {\cellcolor[HTML]{DBDBDB}} 11.7 & 7.2 & 9.9 & 28.8 & 31.5 & 35.1 & 36.9 \\
 & LLAMA\_4 & {\cellcolor[HTML]{DBDBDB}} 11.7 & {\cellcolor[HTML]{FFF4B8}} 12.6 & {\cellcolor[HTML]{F1DBC6}} 11.7 & {\cellcolor[HTML]{DBDBDB}} 14.4 & 33.3 & 33.3 & 25.2 & 25.2 \\
 & MISTRAL & 2.7 & 2.7 & 8.1 & 12.6 & 30.6 & 33.3 & 33.3 & 36.9 \\
\cmidrule{1-10}
\multirow[c]{6}{*}{D, TTL} & GPT\_120 & {\cellcolor[HTML]{FFF4B8}} 12.6 & {\cellcolor[HTML]{FFF4B8}} 12.6 & {\cellcolor[HTML]{DBDBDB}} 12.6 & 12.6 & {\cellcolor[HTML]{FFF4B8}} 38.7 & {\cellcolor[HTML]{FFF4B8}} 40.5 & 30.6 & 31.5 \\
 & GPT\_20 & {\cellcolor[HTML]{FFF4B8}} 12.6 & {\cellcolor[HTML]{FFF4B8}} 12.6 & {\cellcolor[HTML]{DBDBDB}} 12.6 & 12.6 & 27.9 & 30.6 & 24.3 & 25.2 \\
 & GRANITE & 7.2 & 8.1 & 0.9 & 0.9 & 9.9 & 11.7 & 14.4 & 15.3 \\
 & LLAMA\_3 & 6.3 & 9.0 & 8.1 & 12.6 & 36.0 & {\cellcolor[HTML]{FFF4B8}} 40.5 & {\cellcolor[HTML]{FFF4B8}} 40.5 & {\cellcolor[HTML]{FFF4B8}} 45.9 \\
 & LLAMA\_4 & {\cellcolor[HTML]{F1DBC6}} 10.8 & {\cellcolor[HTML]{F1DBC6}} 10.8 & {\cellcolor[HTML]{FFF4B8}} 14.4 & {\cellcolor[HTML]{FFF4B8}} 15.3 & 36.0 & 36.0 & 29.7 & 29.7 \\
 & MISTRAL & 5.4 & 7.2 & 10.8 & 12.6 & {\cellcolor[HTML]{F1DBC6}} 36.9 & {\cellcolor[HTML]{FFF4B8}} 40.5 & 36.0 & {\cellcolor[HTML]{F1DBC6}} 41.4 \\
\cmidrule{1-10}
\multirow[c]{6}{*}{S, BAML} & GPT\_120 & 9.0 & {\cellcolor[HTML]{F1DBC6}} 10.8 & {\cellcolor[HTML]{F1DBC6}} 11.7 & 12.6 & 36.0 & 36.9 & {\cellcolor[HTML]{DBDBDB}} 39.6 & {\cellcolor[HTML]{DBDBDB}} 42.3 \\
 & GPT\_20 & 8.1 & 8.1 & {\cellcolor[HTML]{DBDBDB}} 12.6 & {\cellcolor[HTML]{F1DBC6}} 13.5 & 34.2 & 35.1 & 27.9 & 27.9 \\
 & GRANITE & 1.8 & 6.3 & 3.6 & 7.2 & 18.9 & 19.8 & 21.6 & 27.9 \\
 & LLAMA\_3 & 2.7 & 4.5 & 7.2 & 11.7 & 29.7 & 31.5 & 32.4 & 35.1 \\
 & LLAMA\_4 & 2.7 & 7.2 & 6.3 & {\cellcolor[HTML]{DBDBDB}} 14.4 & 26.1 & 30.6 & 15.3 & 20.7 \\
 & MISTRAL & 6.3 & 6.3 & 5.4 & 6.3 & 36.0 & 37.8 & {\cellcolor[HTML]{F1DBC6}} 36.9 & 39.6 \\
\cmidrule{1-10}
\multirow[c]{6}{*}{S, VERB} & GPT\_120 & \color[HTML]{FFFFFF} nan & \color[HTML]{FFFFFF} nan & \color[HTML]{FFFFFF} nan & \color[HTML]{FFFFFF} nan & {\cellcolor[HTML]{FFF4B8}} 38.7 & {\cellcolor[HTML]{DBDBDB}} 39.6 & \color[HTML]{FFFFFF} nan & \color[HTML]{FFFFFF} nan \\
 & GPT\_20 & \color[HTML]{FFFFFF} nan & \color[HTML]{FFFFFF} nan & \color[HTML]{FFFFFF} nan & \color[HTML]{FFFFFF} nan & 9.9 & 9.9 & \color[HTML]{FFFFFF} nan & \color[HTML]{FFFFFF} nan \\
 & GRANITE & \color[HTML]{FFFFFF} nan & \color[HTML]{FFFFFF} nan & \color[HTML]{FFFFFF} nan & \color[HTML]{FFFFFF} nan & 6.3 & 6.3 & \color[HTML]{FFFFFF} nan & \color[HTML]{FFFFFF} nan \\
 & LLAMA\_3 & \color[HTML]{FFFFFF} nan & \color[HTML]{FFFFFF} nan & \color[HTML]{FFFFFF} nan & \color[HTML]{FFFFFF} nan & 25.2 & 27.9 & \color[HTML]{FFFFFF} nan & \color[HTML]{FFFFFF} nan \\
 & LLAMA\_4 & \color[HTML]{FFFFFF} nan & \color[HTML]{FFFFFF} nan & \color[HTML]{FFFFFF} nan & \color[HTML]{FFFFFF} nan & 29.7 & 30.6 & \color[HTML]{FFFFFF} nan & \color[HTML]{FFFFFF} nan \\
 & MISTRAL & \color[HTML]{FFFFFF} nan & \color[HTML]{FFFFFF} nan & \color[HTML]{FFFFFF} nan & \color[HTML]{FFFFFF} nan & 27.9 & 30.6 & \color[HTML]{FFFFFF} nan & \color[HTML]{FFFFFF} nan \\
\cmidrule{1-10}
\multirow[c]{6}{*}{D, VERB} & GPT\_120 & \color[HTML]{FFFFFF} nan & \color[HTML]{FFFFFF} nan & \color[HTML]{FFFFFF} nan & \color[HTML]{FFFFFF} nan & 31.5 & 32.4 & \color[HTML]{FFFFFF} nan & \color[HTML]{FFFFFF} nan \\
 & GPT\_20 & \color[HTML]{FFFFFF} nan & \color[HTML]{FFFFFF} nan & \color[HTML]{FFFFFF} nan & \color[HTML]{FFFFFF} nan & 23.4 & 27.0 & \color[HTML]{FFFFFF} nan & \color[HTML]{FFFFFF} nan \\
 & GRANITE & \color[HTML]{FFFFFF} nan & \color[HTML]{FFFFFF} nan & \color[HTML]{FFFFFF} nan & \color[HTML]{FFFFFF} nan & 9.0 & 10.8 & \color[HTML]{FFFFFF} nan & \color[HTML]{FFFFFF} nan \\
 & LLAMA\_3 & \color[HTML]{FFFFFF} nan & \color[HTML]{FFFFFF} nan & \color[HTML]{FFFFFF} nan & \color[HTML]{FFFFFF} nan & 33.3 & 34.2 & \color[HTML]{FFFFFF} nan & \color[HTML]{FFFFFF} nan \\
 & LLAMA\_4 & \color[HTML]{FFFFFF} nan & \color[HTML]{FFFFFF} nan & \color[HTML]{FFFFFF} nan & \color[HTML]{FFFFFF} nan & 33.3 & 33.3 & \color[HTML]{FFFFFF} nan & \color[HTML]{FFFFFF} nan \\
 & MISTRAL & \color[HTML]{FFFFFF} nan & \color[HTML]{FFFFFF} nan & \color[HTML]{FFFFFF} nan & \color[HTML]{FFFFFF} nan & {\cellcolor[HTML]{DBDBDB}} 37.8 & {\cellcolor[HTML]{FFF4B8}} 40.5 & \color[HTML]{FFFFFF} nan & \color[HTML]{FFFFFF} nan \\
\cmidrule{1-10}
\bottomrule
\end{tabular}
\end{table}


\begin{table}
\centering
\caption{Percentage of success 
at normalized graph edit distance $NGED^* = 0.2$ 
for all experimental configurations. Each cell reports the score $P_j(0.2)$ 
(see equation \ref{eq:percentage-of-success}) for the configuration as a percentage.
Colors indicate \colorbox[HTML]{ FFF4B8 }{best}, 
\colorbox[HTML]{ DBDBDB }{2nd best}, and \colorbox[HTML]{ F1DBC6 }{3rd best} 
for each column.}
\label{table:ranking-based-on-percentage-of-success-at-nged-0.2}
\begin{tabular}{rr|rrrr|rrrr}
\toprule
 & & \multicolumn{4}{c|}{zero-shot} & \multicolumn{4}{c}{few-shot} \\
\midrule
 & examples & \multicolumn{4}{c|}{False} & \multicolumn{4}{c}{True} \\
 & instances & \multicolumn{2}{c}{False} & \multicolumn{2}{c|}{True} & \multicolumn{2}{c}{False} & \multicolumn{2}{c}{True} \\
 & consolidation & False & True & False & True & False & True & False & True \\
\makecell[r]{instructions\\serialization} & LLM &  &  &  &  &  &  &  &  \\
\midrule
\multirow[c]{6}{*}{S, TTL} & GPT\_120 & 12.6 & 12.6 & 12.6 & 12.6 & {\cellcolor[HTML]{DBDBDB}} 62.2 & {\cellcolor[HTML]{DBDBDB}} 63.1 & 57.7 & 59.5 \\
 & GPT\_20 & 12.6 & 12.6 & 12.6 & 12.6 & 27.9 & 28.8 & 27.9 & 28.8 \\
 & GRANITE & 5.4 & 6.3 & 2.7 & 3.6 & 26.1 & 29.7 & 22.5 & 24.3 \\
 & LLAMA\_3 & 11.7 & 13.5 & 7.2 & 9.9 & 43.2 & 45.0 & 48.6 & 51.4 \\
 & LLAMA\_4 & 14.4 & 17.1 & 20.7 & 24.3 & 45.0 & 44.1 & 36.9 & 37.8 \\
 & MISTRAL & 5.4 & 11.7 & 12.6 & 20.7 & 47.7 & 50.5 & 49.5 & 52.3 \\
\cmidrule{1-10}
\multirow[c]{6}{*}{D, TTL} & GPT\_120 & 12.6 & 12.6 & 12.6 & 12.6 & {\cellcolor[HTML]{DBDBDB}} 62.2 & {\cellcolor[HTML]{DBDBDB}} 63.1 & {\cellcolor[HTML]{F1DBC6}} 58.6 & 61.3 \\
 & GPT\_20 & 12.6 & 12.6 & 12.6 & 12.6 & 50.5 & 55.0 & 47.7 & 52.3 \\
 & GRANITE & 7.2 & 8.1 & 0.9 & 0.9 & 25.2 & 29.7 & 29.7 & 31.5 \\
 & LLAMA\_3 & 8.1 & 11.7 & 11.7 & 18.0 & 50.5 & 51.4 & 55.9 & 62.2 \\
 & LLAMA\_4 & 12.6 & 19.8 & 22.5 & 22.5 & 47.7 & 48.6 & 45.0 & 45.0 \\
 & MISTRAL & 9.0 & 16.2 & {\cellcolor[HTML]{FFF4B8}} 35.1 & {\cellcolor[HTML]{FFF4B8}} 36.0 & {\cellcolor[HTML]{F1DBC6}} 58.6 & 61.3 & {\cellcolor[HTML]{F1DBC6}} 58.6 & 63.1 \\
\cmidrule{1-10}
\multirow[c]{6}{*}{S, BAML} & GPT\_120 & {\cellcolor[HTML]{FFF4B8}} 25.2 & {\cellcolor[HTML]{FFF4B8}} 28.8 & {\cellcolor[HTML]{DBDBDB}} 34.2 & {\cellcolor[HTML]{DBDBDB}} 34.2 & 56.8 & 57.7 & {\cellcolor[HTML]{FFF4B8}} 64.0 & {\cellcolor[HTML]{FFF4B8}} 67.6 \\
 & GPT\_20 & {\cellcolor[HTML]{DBDBDB}} 22.5 & {\cellcolor[HTML]{DBDBDB}} 26.1 & {\cellcolor[HTML]{F1DBC6}} 30.6 & {\cellcolor[HTML]{F1DBC6}} 33.3 & 55.0 & 55.9 & {\cellcolor[HTML]{FFF4B8}} 64.0 & {\cellcolor[HTML]{DBDBDB}} 66.7 \\
 & GRANITE & 9.0 & 13.5 & 7.2 & 10.8 & 30.6 & 33.3 & 35.1 & 46.8 \\
 & LLAMA\_3 & {\cellcolor[HTML]{F1DBC6}} 15.3 & {\cellcolor[HTML]{F1DBC6}} 23.4 & 17.1 & 25.2 & 55.0 & 57.7 & {\cellcolor[HTML]{DBDBDB}} 60.4 & {\cellcolor[HTML]{F1DBC6}} 64.0 \\
 & LLAMA\_4 & 8.1 & 18.9 & 11.7 & 32.4 & 45.0 & 51.4 & 18.9 & 26.1 \\
 & MISTRAL & 11.7 & 15.3 & 14.4 & 15.3 & 55.0 & 59.5 & 54.1 & 57.7 \\
\cmidrule{1-10}
\multirow[c]{6}{*}{S, VERB} & GPT\_120 & \color[HTML]{FFFFFF} nan & \color[HTML]{FFFFFF} nan & \color[HTML]{FFFFFF} nan & \color[HTML]{FFFFFF} nan & {\cellcolor[HTML]{FFF4B8}} 67.6 & {\cellcolor[HTML]{FFF4B8}} 67.6 & \color[HTML]{FFFFFF} nan & \color[HTML]{FFFFFF} nan \\
 & GPT\_20 & \color[HTML]{FFFFFF} nan & \color[HTML]{FFFFFF} nan & \color[HTML]{FFFFFF} nan & \color[HTML]{FFFFFF} nan & 18.9 & 22.5 & \color[HTML]{FFFFFF} nan & \color[HTML]{FFFFFF} nan \\
 & GRANITE & \color[HTML]{FFFFFF} nan & \color[HTML]{FFFFFF} nan & \color[HTML]{FFFFFF} nan & \color[HTML]{FFFFFF} nan & 19.8 & 21.6 & \color[HTML]{FFFFFF} nan & \color[HTML]{FFFFFF} nan \\
 & LLAMA\_3 & \color[HTML]{FFFFFF} nan & \color[HTML]{FFFFFF} nan & \color[HTML]{FFFFFF} nan & \color[HTML]{FFFFFF} nan & 39.6 & 45.0 & \color[HTML]{FFFFFF} nan & \color[HTML]{FFFFFF} nan \\
 & LLAMA\_4 & \color[HTML]{FFFFFF} nan & \color[HTML]{FFFFFF} nan & \color[HTML]{FFFFFF} nan & \color[HTML]{FFFFFF} nan & 43.2 & 44.1 & \color[HTML]{FFFFFF} nan & \color[HTML]{FFFFFF} nan \\
 & MISTRAL & \color[HTML]{FFFFFF} nan & \color[HTML]{FFFFFF} nan & \color[HTML]{FFFFFF} nan & \color[HTML]{FFFFFF} nan & 46.8 & 48.6 & \color[HTML]{FFFFFF} nan & \color[HTML]{FFFFFF} nan \\
\cmidrule{1-10}
\multirow[c]{6}{*}{D, VERB} & GPT\_120 & \color[HTML]{FFFFFF} nan & \color[HTML]{FFFFFF} nan & \color[HTML]{FFFFFF} nan & \color[HTML]{FFFFFF} nan & 57.7 & {\cellcolor[HTML]{F1DBC6}} 62.2 & \color[HTML]{FFFFFF} nan & \color[HTML]{FFFFFF} nan \\
 & GPT\_20 & \color[HTML]{FFFFFF} nan & \color[HTML]{FFFFFF} nan & \color[HTML]{FFFFFF} nan & \color[HTML]{FFFFFF} nan & 45.9 & 50.5 & \color[HTML]{FFFFFF} nan & \color[HTML]{FFFFFF} nan \\
 & GRANITE & \color[HTML]{FFFFFF} nan & \color[HTML]{FFFFFF} nan & \color[HTML]{FFFFFF} nan & \color[HTML]{FFFFFF} nan & 23.4 & 27.0 & \color[HTML]{FFFFFF} nan & \color[HTML]{FFFFFF} nan \\
 & LLAMA\_3 & \color[HTML]{FFFFFF} nan & \color[HTML]{FFFFFF} nan & \color[HTML]{FFFFFF} nan & \color[HTML]{FFFFFF} nan & 49.5 & 52.3 & \color[HTML]{FFFFFF} nan & \color[HTML]{FFFFFF} nan \\
 & LLAMA\_4 & \color[HTML]{FFFFFF} nan & \color[HTML]{FFFFFF} nan & \color[HTML]{FFFFFF} nan & \color[HTML]{FFFFFF} nan & 45.9 & 46.8 & \color[HTML]{FFFFFF} nan & \color[HTML]{FFFFFF} nan \\
 & MISTRAL & \color[HTML]{FFFFFF} nan & \color[HTML]{FFFFFF} nan & \color[HTML]{FFFFFF} nan & \color[HTML]{FFFFFF} nan & {\cellcolor[HTML]{F1DBC6}} 58.6 & 61.3 & \color[HTML]{FFFFFF} nan & \color[HTML]{FFFFFF} nan \\
\cmidrule{1-10}
\bottomrule
\end{tabular}
\end{table}

At a threshold of \( NGED^* = 0.1 \), as detailed in table \ref{table:ranking-based-on-percentage-of-success-at-nged-0.1}, configurations that serialize the ontology schema into TTL format achieve the highest percentages of success. In zero-shot settings, the most effective configuration scores 15.3\%, using the \text{LLAMA\_4} model with detailed instructions, along with TTL serialization, instances, and consolidation. In few-shot scenarios, the best result is a 45.9\% percentage of success, obtained by employing the same configuration but with the \text{LLAMA\_3} model. As expected, few-shot configurations substantially outperform zero-shot ones. Specifically, the top performance in few-shot settings (45.9\%) represents a 200\% improvement compared to the best score in zero-shot settings (15.3\%).

When we increase the threshold \( NGED^* \) to 0.2 (table \ref{table:ranking-based-on-percentage-of-success-at-nged-0.2}), which corresponds to a higher tolerance for prediction errors, both zero-shot and few-shot settings achieve an increased percentage of success. In the zero-shot setting, the best performance (36.0\%) is still achieved by using detailed instructions along with TTL serialization, instances, and consolidation. However, this time, the \text{MISTRAL} model outperforms the previously used \text{LLAMA\_4} model. Interestingly, in few-shot settings, the \text{GPT\_120} model achieves notable results (67.6\%) when combined with three different configurations: one using simple instructions, BAML serialization of the schema, instances, and consolidation; two variations employing simple instructions and VERB format for schema serialization without instances. Interestingly, the smaller \text{GPT\_20} model achieves a similar result (66.7\%), when combined with simple instructions, BAML serialization, instances, and consolidation. Overall, at \( NGED^* = 0.2 \), BAML serialization appears to enhance zero-shot and few-shot configurations more effectively than TTL serialization does. Furthermore, we observe that few-shot settings continue to outperform zero-shot ones under this threshold, with the top few-shot result (67.6\%) representing an 87.8\% improvement over the best zero-shot score (36.0\%).

To gain deeper insight into our experimental results and patterns seen on the table, we conducted statistical analyses to compare families of configurations. 
A family $\mathcal{F} = \left\{ c_1, c_2, \ldots, c_m \right\}$ consists of a set of configurations that share some common prompt features: the format of instructions, either simple (S) or detailed (D), the schema serialization method (TTL or BAML), and whether they use examples, instances or consolidation. Due to the intrinsic limitations of the VERB schema serialization, only a few configurations utilize this approach. To ensure balanced comparisons among families of configurations, we exclude those employing VERB serialization from our analysis.

In our comparison of configuration families, we analyze their respective distributions of percentages of success across various threshold values \( NGED^* \): 0.1, 0.2, 0.3, 0.4, and 0.5. We do not extend this analysis beyond \( NGED^* = 0.5 \), because higher thresholds would result in extraction outcomes with an excessive number of errors (manifesting as KGs with too many incorrect nodes or edges). The elements of the array \( P_{\mathcal{F}}(NGED^*) \) represent the percentages of success at threshold $NGED^*$ for configurations belonging to family \( \mathcal{F} \).

When comparing two distributions, we use the Mann-Whitney U test \cite{mann1947}; when comparing three or more distributions, we use the Kruskal-Wallis test \cite{kruskal1952use} followed by post hoc analysis with Dunn's test \cite{c3fa9fa7-dd2a-35f2-84a0-d5c07e68dd08} with Holm's correction \cite{holm1979simple}. We use the function \texttt{mannwhitneyu}, from the SciPy library \cite{2020SciPy-NMeth}, with parameter \texttt{method='exact'} to perform the Mann-Whitney U test; we use the \texttt{kruskal} function from the SciPy library to perform the Kruskal-Wallis test, and the \texttt{posthoc\_dunn} function from the scikit-posthocs library \cite{Terpilowski2019} to perform the post hoc analysis with the Dunn's test with Holm's correction.

\begin{figure}
    \centering
    \includegraphics[width=\linewidth]{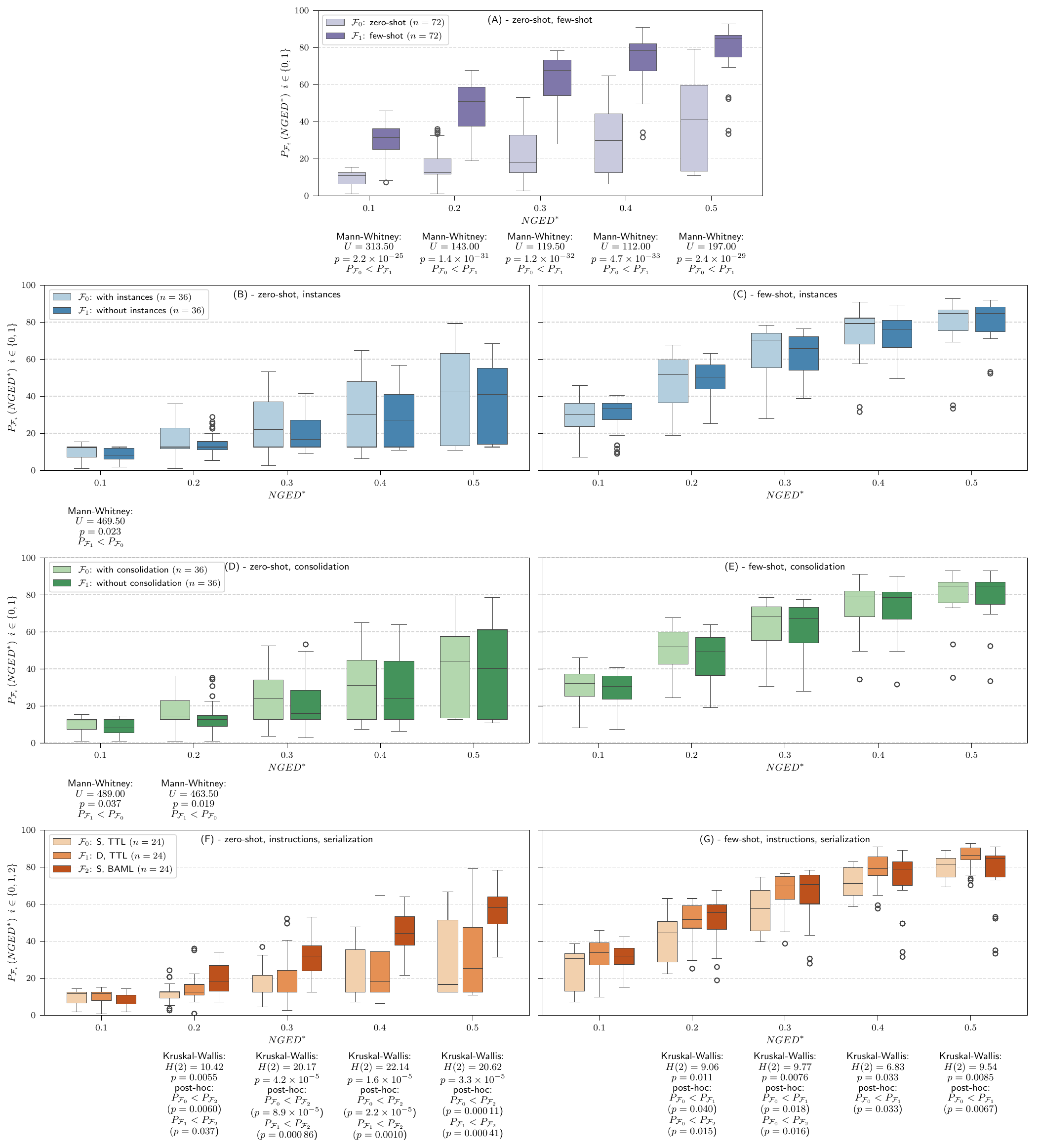}
    \caption{Percentages of success \( P_{\mathcal{F}} \) at various thresholds of normalized graph edit distance (\( NGED^* \)) across different configuration families \(\mathcal{F}\) ($n = \ldots $ is the number of configuration within the family). When significant results are found, we report the outcomes of statistical tests beneath each box plot. For comparisons between two families, we use the Mann-Whitney U test, otherwise we use the Kruskal-Wallis test followed by pairwise post hoc analysis using Dunn's test with Holm's correction. We conduct all possible post hoc pairwise comparisons, but we report only those that present statistically significant differences: \( P_{\mathcal{F}_i} < P_{\mathcal{F}_j} \) indicates that the distribution of percentages of success for \(\mathcal{F}_i\) is statistically significantly lower than that for \(\mathcal{F}_j\).}
    \label{fig:statistical_analysis}
\end{figure}

Figure \ref{fig:statistical_analysis}(A) compares the family $\mathcal{F}_0$, comprising 72 zero-shot configurations, with the family $\mathcal{F}_1$, comprising 72 few-shot configurations. Our analysis revealed that $P_{\mathcal{F}_0}$ is statistically significantly lower than $P_{\mathcal{F}_1}$ for all values of $NGED^*$. This finding confirms that in-context learning with few-shot configurations generally outperforms zero-shot configurations. Moreover, we observe an increase in the median percentages of success for both $\mathcal{F}_0$ (zero-shot) and $\mathcal{F}_1$ (few-shot) as $NGED^*$ increases. This trend is expected because raising the threshold value of $NGED^*$ implies a greater tolerance to errors, resulting in higher percentages of success. The plot also shows a wider interquartile range (IQR) for $P_{\mathcal{F}_0}$ compared to $P_{\mathcal{F}_1}$ as $NGED^*$ rises. This indicates increased variability of the percentages of success for zero-shot configurations, suggesting that  additional factors related to prompt structure (beyond just inclusion of examples) might influence performance, especially in a zero-shot context.

To further enhance our understanding, we conducted separate investigations into how various features of prompts (such as the inclusion of instances, the use of consolidation heuristics, varied instructions formats, and different serialization techniques for the ontology schema) affect the percentages of success in zero-shot and few-shot configurations. 

A Mann-Whitney U test revealed no statistically significant differences between the distributions of percentages of success for configuration families with prompts containing instances versus those without, both in zero-shot (figure \ref{fig:statistical_analysis}(B)) and few-shot (figure \ref{fig:statistical_analysis}(C)) scenarios. This suggests that including instances in the prompt does not meaningfully impact performance outcomes between these groups. The only noticeable exception is for zero-shot configurations with threshold $NGED^* = 0.1$, where the absence of instances both in the prompt and in the examples, yields configurations whose percentages of success are statistically significantly lower compared to those configurations using instances. Similarly, the use of consolidation heuristics significantly affects only the percentage of success of zero-shot configuration families at $NGED^*$ threshold of 0.1 and 0.2, as shown in figure \ref{fig:statistical_analysis}(D). However, these effects are not observed at higher values of \( NGED^* \), nor for few-shot configurations (Figure \ref{fig:statistical_analysis}(E)). This pattern suggests that consolidation heuristics become less effective for zero-shot configuration when the chosen metric tolerates more extraction errors (at higher values of $NGED^*$), or when the prompt already includes examples.

Finally, we explored how the complexity of instructions—whether simple (S) or detailed (D)—and schema serialization methods (TTL or BAML) influence the percentage of success. Figures \ref{fig:statistical_analysis}(F) and \ref{fig:statistical_analysis}(G) illustrate our findings for zero-shot and few-shot configurations, respectively. At \( NGED^* = 0.1 \), a Kruskal-Wallis test shows no significant differences, regardless of whether examples are included in the prompt. This suggests that when the metric is very stringent (indicating low tolerance for extraction errors), variations in instruction or serialization formats do not significantly impact performance outcomes. In contrast, we find statistically significant differences when $NGED^* \ge 0.2$. When examples are included in the prompt along with TTL serialization, configurations using simple instructions result in percentage of success distributions that are statistically significantly lower compared to those using detailed instructions. In certain cases ($NGED^*$ of 0.1 or 0.2), these distributions are also lower than those obtained by configurations combining simple instructions with BAML serialization. Noticeably, in zero-shot configurations with \( NGED^* \ge 0.2 \), the statistical test shows that using TTL serialization leads to percentage of success distributions that are statistically significantly lower than those achieved with BAML serialization, regardless of whether simple or detailed instructions are used. Additionally, in the zero-shot setting, the interquartile range for configuration families employing BAML serialization does not widen as much as it does for those
using TTL when \( NGED^* \) increases. These observations suggest that BAML serialization is a distinguishing factor contributing to better results in scenarios where examples cannot be included in LLM prompts.

\subsection*{Guidelines}

Our findings demonstrate that LLMs are capable of generating knowledge graphs that are both syntactically and semantically valid according to a specified ontology, even in the absence of domain-specific fine-tuning. Through our experiments, we have identified practical guidelines to effectively extract ontology-compliant KGs from text.

\begin{itemize}

    \item \textbf{Generating examples for few-shot configurations using zero-shot techniques.} In zero-shot settings, LLMs can generate syntactically valid KGs. However, ensuring these KGs comply semantically with a given ontology is challenging. High-quality examples of KGs are not always available, making it difficult to guide the model effectively. To address this, we recommend selecting representative and varied text passages, and using them as inputs for a zero-shot configuration that includes simple instructions, the ontology schema serialized in BAML format, available instances (if any), and consolidation heuristics. This method helps to extract initial KGs with reasonable accuracy, which can then serve as examples for few-shot configurations.

    \item \textbf{Validation by domain experts.} We recommend involving domain experts in reviewing KGs generated through zero-shot configurations to identify and correct any inaccuracies. By iteratively refining these KGs, a set of high-quality examples can be developed. This process significantly enhances the effectiveness of subsequent extractions using the validated KGs in few-shot configurations.

    \item \textbf{Utilizing few-shot configurations when high-quality examples are available.} When high-quality examples are accessible, employing few-shot configurations is highly recommended. Our experiments demonstrate that few-shot approaches significantly outperform zero-shot ones. To optimize results, we recommend to include 10-15 high-quality examples in the prompt, along with detailed instructions and TTL serialization of the ontology.

    \item \textbf{Selecting appropriate serialization and instruction formats.} When serializing the ontology in TTL format, configurations with detailed instructions consistently outperform those with simple ones across various LLMs. This holds true even if the instructions are not specifically tailored to a particular model, highlighting the significance of explicit task guidance. In contrast, when using BAML serialization, which yields better results in zero-shot settings, simple instructions are sufficient.

    \item \textbf{Usefulness of verbalized serialization.} Models generally interpret complete serialized ontologies (either in TTL or BAML format) more effectively than simple verbalizations, as they can understand semantic constraints like domains and ranges of properties. The VERB configuration, though insufficient for zero-shot configurations due to its lack of explicit constraint information, remains useful in few-shot scenarios, for example when the ontology is only partially defined (listing just classes and their relations).

    \item \textbf{Opting for few-shot prompts with high-quality examples over instances.} Incorporating ontology instances generally has little impact on performance, except in zero-shot configurations at very strict $NGED^*$ thresholds. While instances can aid in mapping lexical mentions to the correct entities when these are absent from examples, few-shot prompts that include only the schema often perform better. This suggests that high-quality examples provide stronger inductive signals than lists of instances. Examples typically illustrate connected subgraphs with instantiated relations, offering structural context that instance dictionaries lack. Instead of adding lengthy or incomplete list of instances to prompts, a lightweight post-processing step can efficiently enhance instance mapping.

    \item \textbf{Effectiveness of ontology-guided consolidation with simple heuristics}. Ontology-guided consolidation provides cost-effective improvements using straightforward, domain-agnostic heuristics. This is particularly evident with smaller models and zero-shot configurations at strict thresholds (\(NGED^{*} \le 0.2\)). The impact of consolidation heuristics diminishes when examples are included in prompts. Although post-processing can resolve surface-level inconsistencies, in-context learning ensures better semantic correctness.
    
\end{itemize}

\section{Discussion}

The study and discovery of materials capable of withstanding extreme temperatures and radiation levels in fusion reactors are hindered by the lack of structured, machine-readable data. Much critical information is embedded within dense scientific literature, making manual extraction time-consuming. Domain experts typically need 30 to 90 minutes per article to distill relevant data, slowing innovation and limiting access to valuable insights. In contrast, our \textit{eolas} pipeline efficiently generates knowledge graphs from hundreds of articles in just a few hours. Additionally, domain experts can quickly navigate and validate the extracted information using a simple user interface, enhancing both speed and accessibility.

We introduce a novel ontology designed to capture the complex semantic relationships of defects in irradiated materials. Building upon this ontology, we present the first comprehensive benchmark dataset specifically created for evaluating LLMs in their ability to extract knowledge graphs that adhere to the given ontology from scientific texts. Our work highlights the potential of KGs as interpretable representations that facilitate AI-driven scientific exploration. The International Atomic Energy Agency (IAEA) has acknowledged this transformative approach \cite{international2021iaea}, advocating for semantic technologies to manage distributed nuclear knowledge, enhance search and retrieval capabilities, support automated reasoning, and foster novel data-driven discoveries..

Utilizing \textit{eolas}, our modular extraction pipeline, and benchmark dataset, we conducted 168 experiments using various prompting techniques and publicly available LLMs (table \ref{table:experiments_summary}). These experiments resulted in the evaluation of 18,648 knowledge graphs. Unlike prior studies such as \cite{Dagdelen2024}, \cite{Polak_2024} and \cite{dasilva2024automatedllmenabledextraction}, which focus on using LLMs to extract JSON data with relatively simple schemas, our approach builds knowledge graphs that align with a comprehensive ontology. We found that the formal axioms of an ontology can enhance LLM-extracted data by identifying implausible values, enforcing constraints, or inferring contextual information. Additionally, previous research does not specifically target the domain of irradiated materials nor provide an openly accessible benchmark dataset for systematic evaluation.

The versatility and modularity of \textit{eolas} allowed us to efficiently evaluate several configurations across various parameters. These include testing six different LLMs, two instruction formats, three ontology serialization formats, the use of instances and examples, as well as consolidation heuristics. This flexibility sets our approach apart from existing methods. Our study is unique in offering an extensive and systematic evaluation of LLM capabilities to extract knowledge graphs that conform to a specified schema from text. Our extensive experiments provide valuable guidance for scientists navigating the complex process of accurately extracting knowledge from text with LLMs.

While our extraction pipeline is designed to be domain-independent, this study specifically focuses on extracting knowledge graphs related to irradiated materials —a challenging domain characterized by complex information and specialized terminology. Future research could explore the applicability of \textit{eolas} across different domains and evaluate the utility of the guidelines we offer. Additionally, we observe that we adopt a strict metric in evaluating the results of our experiments: when computing the (normalized) graph edit distance, we compare values (nodes or edges labels) using equality. This approach penalizes situations in which the LLM extracts approximations or synonyms instead of the expected values. Similarly, an LLM may introduce additional correct information (nodes or edges in the graph) that is not originally present in the text, and therefore absent from the benchmark graphs: in these cases the graph edit distance metric penalizes the LLM, even when the added content is factually correct. Future  research could integrate heuristics and approximate matching techniques when computing the graph edit distance. This approach might yield more insightful evaluation results, even if they are less stringent. Additionally, employing alternative methods such as LLM-as-a-judge \cite{zheng2023judging, liu-etal-2023-g, chiang-lee-2023-large} could offer valuable complementary evaluations for our approach.

Additional challenges remain in merging knowledge graphs extracted at paragraphs level into a coherent knowledge graph at document level, or across documents. A promising future research direction to tackle this problem is the use of an interactive, LLM-based, multi-agent system, in which multiple agents may collaborate, revise, and improve each other work, while dynamically interacting with a human expert (similar to the conversational approach adopted in \texttt{ChatExtract} \cite{Polak_2024} to allow an LLM revise its responses). 

Despite some limitations, our work has immediate practical applications. The knowledge graphs we extract provide the structured data necessary for setting up atomistic simulations (Kinetic Monte Carlo or Cluster Dynamics)
by providing required parameters, such as migration energies, binding energies or defect configurations. Using these knowledge graphs, researchers can streamline high-throughput studies and reduce human error during simulation setup. 

Furthermore, the knowledge graphs we have extracted can be directly integrated into advanced information retrieval systems like Graph-RAG \cite{edge2024from, wang2024graph}. This integration enables sophisticated, accurate, and domain-specific question answering. To validate this, we conducted an experiment using a simplified version of Graph-RAG based on our ontology and some competency questions defined by domain experts. The results confirmed that the ontology significantly enhanced the LLM's ability to provide accurate answers (see details in Appendix \ref{secA2}).

In summary, we found that LLMs are capable of effectively interpreting an ontology schema to extract compliant knowledge graphs from text, achieving impressive results in both few-shot (67.6\% percentage of success at \(NGED^* = 0.2\)) and zero-shot scenarios (36.0\% percentage of success at \(NGED^* = 0.2\)). Importantly, our analysis of numerous experiments has yielded general guidelines to assist future efforts focused on extracting knowledge graphs from text using LLMs and input schemas.

\section{Methods}

\subsection{Ontology/Schema building}

The ontology was developed in collaboration with a multidisciplinary team of domain experts to define key concepts using meaningful labels and, optionally, short descriptions. It also specifies semantic relationships along with constraints on domains and ranges. The ontology is designed to accurately and unambiguously represent the domain of irradiated materials by capturing knowledge from both simulations and experimental studies related to atomistic defect energetics in materials. We iteratively refined the ontology using the Protégé ontology editing environment \cite{protege}. As an evolving resource, the ontology can expand its coverage of crystallographic defects by aligning with other upper-level materials otologies, such as the European Materials Modelling Ontology (EMMO) \cite{emmo}. Importantly, our knowledge extraction pipeline
remains schema-agnostic, facilitating seamless integration with various ontological frameworks as needed.

\begin{figure}
\centering
\includegraphics[width=\textwidth]{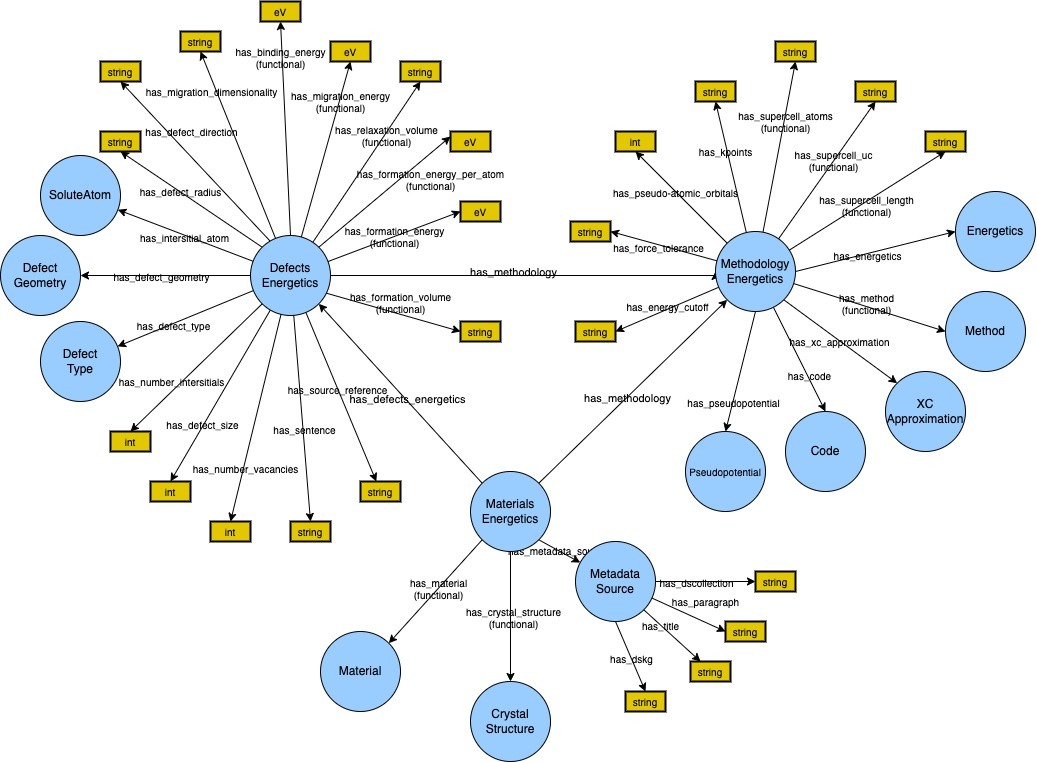}
\caption{Structure of the proposed ontology to model the domain of irradiated materials}
\label{figonto}
\end{figure}

Several best practices exist for manual ontology development \cite{noy2001ontology, methontology}. The core process involves defining the ontology's purpose and scope, followed by collaboration between ontology engineers and domain experts to identify key conceptual elements.
Our development process began with domain experts manually identifying key examples from representative papers \cite{multiscale-modelling-irradiated, relaxation, PhysRevB.92.104102, barouh_PhysRevB.90.054112, MALERBA2021_parameters}. These examples were used to populate a table describing relevant properties and values, which then informed the definition of ontology classes, object and data properties, instances, and associated semantic constraints.
At its core, as illustrated in figure~\ref{figonto}, a root node (\texttt{MaterialsEnergetics}) connects all other nodes, either directly or through intermediary paths. A property can have multiple domains; for example, \texttt{has\_methodology} links the range \texttt{MethodologyEnergetics} with the domains \texttt{MaterialEnergetics} and \texttt{DefectsEnergetics}. This link is relevant when methodologies are reported for specific defect type measurements.
We distinguish between intermediate and leaf nodes. Intermediate nodes, such as \texttt{MaterialsEnergetics}, \texttt{DefectsEnergetics}, and \texttt{MethodologyEnergetics}, do not have meaningful labels themselves but serve as structural components to model complex relationships, akin to reification in RDF/OWL \cite{cyganiak2014rdfs11}.
For instance, an entity of type \texttt{DefectsEnergetics} can describe properties and attributes for a given defect, such as relaxation volume, migration energies, binding energy, defect geometry, size, solute atom presence, etc. Intermediate properties connect these intermediate nodes, while leaf properties are either object or datatype properties that link an intermediate node to a leaf node or literal (such as string or numerical values like defect size or relaxation volume, which may be
expressed as a number or range). Leaf properties are uniquely labeled to ensure consistency when automatically creating prompts. Additionally, functional constraints can be applied to leaf properties to enforce unique values for each instance.

Leaf properties and leaf nodes are essential for transforming knowledge graphs into a user-friendly tabular format. In this format, leaf properties correspond to table columns, while leaf nodes provide the respective values. The property-values of connected intermediate nodes are displayed in the same row.
Optionally, dictionaries can be defined to capture a non-exhaustive set of known relevant instances for a given leaf node in the ontology. These dictionaries are stored as JSON files and automatically integrated into a populated ontology as predefined instances with a preferred label and a list of alternative labels or synonyms. A class in the ontology is linked to a dictionary using the property \texttt{rdfs:isDefinedBy}.
This approach supports the development of a common vocabulary, which helps in constructing a document-level KG and facilitates end-users in retrieving relevant information across various sources.

Competency questions are often employed to assess whether an ontology can adequately represent the knowledge required to fulfill its intended purpose - by verifying it can answer those questions. In our case, the ontology was refined primarily to capture the relevant contextual information needed during ground-truth construction.
A set of expert-defined competency questions, along with their corresponding answers generated using a knowledge graph derived from our ground truth are provided in Appendix~\ref{secA2}. These examples illustrate the value of structured knowledge in answering multi-hop queries that span multiple paragraphs or documents.

\subsection{\textit{eolas} extraction pipeline}

We have developed \textit{eolas} as a modular pipeline that can be easily extended and customized. The core functionality includes several key steps: pre-processing PDF documents to prepare them for analysis; extracting knowledge graphs from text fragments within each document, guided by the input ontology schema; consolidating these partial graphs into comprehensive, document-level knowledge graphs.

\subsubsection{Step 1 - document ingestion and pre-processing}

We use Docling \cite{Docling, Docling_Team_Docling} for PDF documents conversion, and Deep Search \cite{deepsearch, Auer_2022} to build scalable, custom and searchable collections of documents (e.g., the full arXiv corpus or a subset of it focused on relevant materials such as iron, iron alloys or tungsten). The extraction can handle both text-based and image-based PDFs through integrated OCR capabilities, and parses document level structures such as sentences, paragraphs, tables, figures, headers, captions, and images. Deep Search also provides domain-specific dictionary-based annotators that categorize and label text.

\subsubsection{Step 2 - filtering}

After parsing the documents, we obtain both paragraphs and sentences along with annotations made by the dictionary-based Deep Search annotators. At this stage, we can filter out irrelevant passages—whether they are paragraphs or sentences—using configurable, schema-driven heuristics. For instance, we might discard sentences or paragraphs that have few or no annotations, or those that are citations. This optional, lightweight filtering step helps reduce processing costs by limiting the number of passages sent to a large language model in the following steps.

\subsubsection{Step 3 - annotation}

In some scenarios, Deep Search annotations may be sufficient to filter out irrelevant passages. However, certain cases may require more complex annotators due to their complexity or specificity. Additionally, there may be concepts in the ontology or schema that are not fully covered by existing Deep Search annotators or dictionaries of instances. 
As an optional step, our pipeline supports the use of alternative annotators—such as LLM-based annotators \cite{pavlovic-poesio-2024-effectiveness} or libraries for zero-shot entity recognition \cite{picco-etal-2023-zshot}. These alternatives can enhance entity coverage and ensure better alignment with the ontology. This optional step not only aids in passage filtering and hallucination detection but also facilitates entity matching. Moreover, it enables precise highlighting of relevant text or entities in the user interface, which is a useful feature for end users (see figure \ref{fig:diagrams-eolas_user_interface_with_graph}(b)).

\subsubsection{Step 4 - merging annotations}

This step consolidates all annotations generated by Deep Search, along with any additional annotations from optional tools introduced in Step 3. It eliminates duplicate entries and resolves ambiguities related to entities. Optionally, the pipeline can leverage these consolidated annotations to further filter text passages, and reduce the number of texts sent to a large language model for knowledge graph extraction, thus enhancing efficiency and lowering processing costs.

\subsubsection{Step 5 - knowledge graph extraction}

This is the main step in our pipeline and involves an extractor component that interfaces with a large language model to extract knowledge graphs from text fragments based on a specified ontology schema. The configuration of the extractor defines how LLM prompts are constructed, detailing elements such as the level of instructions (simple or detailed), the serialization format for the ontology schema (TTL, BAML, VERB), and whether instances and examples should be included. This step ensures modularity and adaptability of the extraction process.

\begin{figure}
    \centering
    \includegraphics[width=\linewidth]{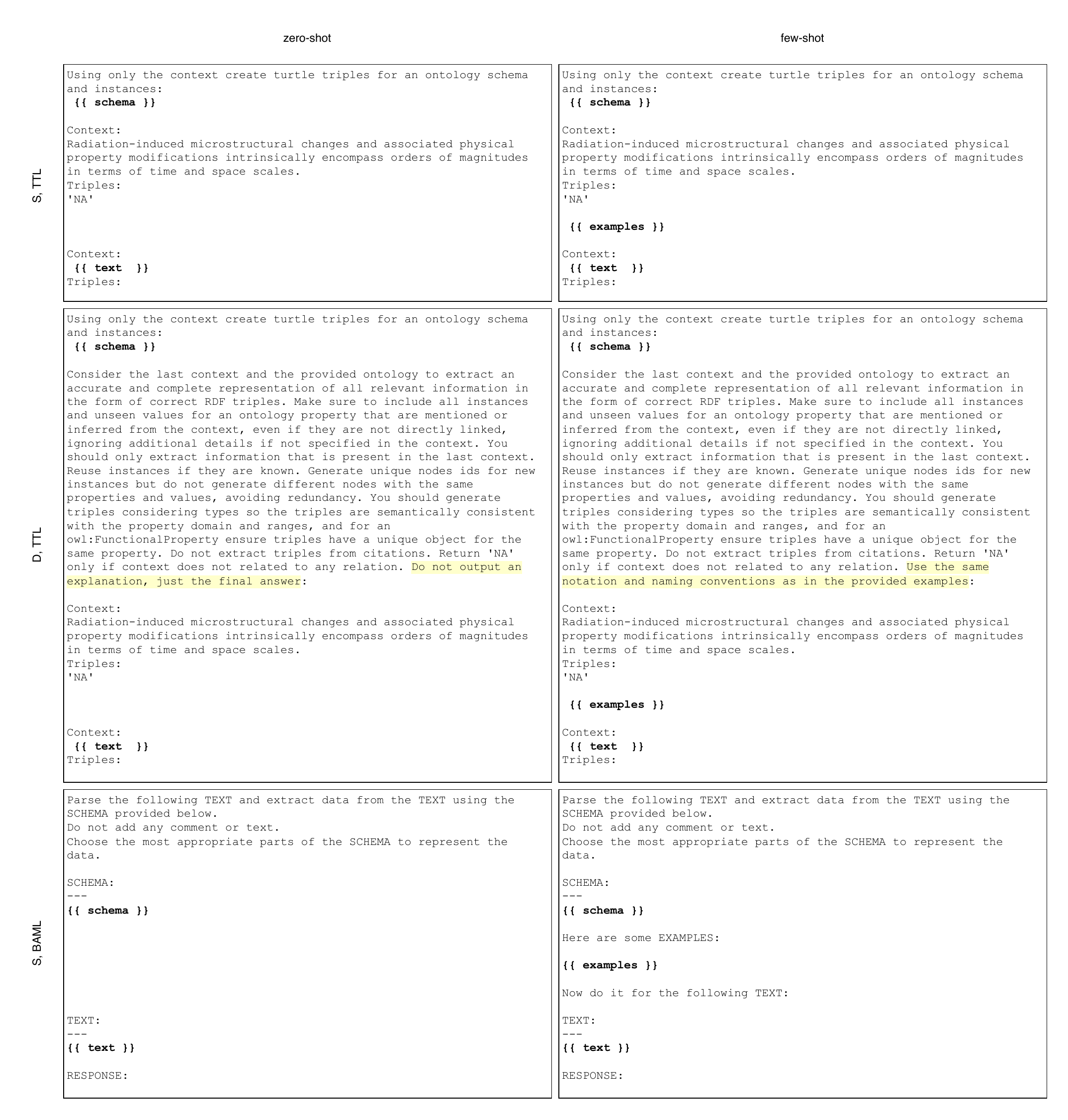}
    \caption{Templates for prompt formats used in zero-shot configurations (on the left) and few-shot configurations (on the right). The first two rows display templates utilized when serializing the ontology schema in TTL format with simple instructions and detailed instructions, respectively; there is a small difference (highlighted in yellow) between the detailed instructions for few-shot and zero-shot templates: the former refer to the examples, while the latter do not. The third row presents the template utilized when serializing the ontology schema in BAML format with simple instructions. The templates for the VERB serialization format are the same as those for S, TTL and D, TTL. Sections highlighted in bold (denoted by \texttt{\{\{ schema \}\}}, \texttt{\{\{ examples \}\}}, and \texttt{\{\{ text \}\}}) are placeholders, and are replaced at runtime with appropriate content (the serialized ontology schema, the examples, and the input text to convert into a knowledge graph). Note that empty lines are added for readability purposes in the figure but do not necessarily form part of the actual prompt.}
    \label{fig:prompts.drawio}
\end{figure}

Figure \ref{fig:prompts.drawio} illustrates the various templates employed to dynamically generate prompts tailored to various configurations. The prompt templates used with TTL serialization include by default a negative example to show the LLM that not every passage contains a relevant knowledge graph. In such cases, the model is instructed to generate 'NA'. 

We currently have three specialized implementations of the extractor component, each designed for a specific serialization format: TTL (turtle triples), verbalized format (VERB), and BAML.

\begin{description}
    \item[TTL Extractor.] This implementation serializes the ontology schema using turtle triples. If instances are part of the configuration, they too are serialized as turtle triples immediately following the schema. For few-shot configurations, a KG is provided containing examples compliant with the ontology schema and aligned with the source text. Each example is automatically incorporated into the prompt in two sections: a "Context:" section containing the text passage, and a "Triples:" section presenting the corresponding knowledge graph, also serialized as turtle triples. This extractor expects the LLM to produce as output valid and connected turtle triples, which form the predicted knowledge graph.

    \item[VERB Extractor.] This implementation transforms the ontology schema in a textual format (verbalization), which comprises a list of classes and their relations (i.e., those object and datatype properties for which the class is the domain); the verbalization contains only the labels of the ontology classes and relations. This extractor does not serialize instances, and it handles examples (few-shot) using the same format as the TTL Extractor. Also, this extractor expects the LLM to produce as output valid turtle triples.

    \item[BAML Extractor.] This extractor converts the OWL \cite{OWL} ontology into BAML \cite{Boundary_BAML_-_A_2025} format. During the conversion we preserve labels (compatibly with syntax limitations), and we convert OWL descriptions into BAML comments. We convert each OWL class $\mathbf{C}^O_i$ into a BAML class $\mathbf{C}^B_i$. An OWL datatype property having domain $\mathbf{C}^O_i$, is converted to a field of $\mathbf{C}^B_i$ with the same range (BAML supports \texttt{string}, \texttt{int}, \texttt{float}, and \texttt{bool}). An OWL object property, having domain $\mathbf{C}^O_i$ and range $\mathbf{C}^O_j$, is converted to a field of $\mathbf{C}^B_i$ with range $\mathbf{C}^B_j$. BAML supports constructs such as \texttt{list}, \texttt{optional}, and \texttt{union}, which facilitate modeling common OWL features, including class unions or functional properties. If the ontology includes instances of classes (specified with \texttt{owl:NamedIndividual}), these are converted into members of a BAML \texttt{Enum}, an enumeration of constant values (possibly with descriptions). The BAML approach guides the LLM to generate JSON output that aligns with BAML class definitions. Once this JSON is produced, our extractor automatically transforms it into turtle triples, which are structured as a knowledge graph consistent with the ontology schema. In parallel, when using few-shot prompts, the BAML Extractor takes the knowledge graphs corresponding to examples included in the prompt, and converts them into JSON data structures. These structures comply with the BAML serialization of the ontology, ensuring consistency across both input examples and extracted outputs.
\end{description}

The architecture of our extraction pipeline is designed to be modular, allowing for seamless integration of additional extractors tailored to various configurations.

The extractor component generates a prompt for each text fragment that requires conversion into a knowledge graph. Several factors influence the size of these prompts (measured in tokens), including the length of the text fragment, the ontology's size—which varies with the chosen serialization format —and whether instances or examples are included. Although different LLMs have context window of varying sizes, it is crucial to consider the text fragment's size for knowledge graph generation. Longer passages of text, such as paragraphs, provide more contextual information but can also challenge LLMs in managing multiple entities and their complex relationships. This complexity might lead to increased errors in the predicted knowledge graphs. Conversely, shorter fragments, like single sentences, may lack sufficient context, resulting in fragmentation errors. We conducted some preliminary experiments with the set of LLMs used in our study (see table \ref{table:experiments_summary}), and we found that extracting at the paragraph level generally yields better results than at the sentence level. Consequently, we divide input text into paragraphs for processing. However, alternative strategies might be more effective when dealing with documents containing unusually long paragraphs.

The content generated by the LLM for each segment of text is incorporated into a graph (we use the Pyhton library RDFLib \cite{Krech_RDFLib_2025}); if the LLM output does not comply with syntactic standards of RDF/Turtle format, then the construction of the graph will be unsuccessful.

\subsubsection{Step 6 - consolidation heuristics}

This optional consolidation step of the extraction pipeline consists of a set of semantic heuristics designed to enhance both the quality and consistency of the knowledge graphs generated by the LLM.

\begin{description}
    \item[Consolidation of nodes identifiers.] LLMs may not consistently extract references to the same entity using identical identifiers, which can lead to duplicated entity nodes in the knowledge graph. To address this issue, we employ fuzzy matching techniques using the RapidFuzz library \cite{rapidfuzz}. This approach attempts to match entities with unknown identifiers against known instances. If a close match is identified through this process, then the existing identifier of the matched entity is reused, otherwise a new identifier is assigned to ensure each entity remains distinct and properly referenced in the knowledge graph.

    \item[Detection and reduction of hallucinations.] LLMs sometimes generate plausible yet incorrect information, a phenomenon known as "hallucination" \cite{kalai2025languagemodelshallucinate}. In our specific application, hallucinations occur when an LLM produces triples in the predicted knowledge graph that lack corresponding evidence in the input text. Although some of these generated triples may be factually correct—due to the LLM's ability to leverage domain knowledge acquired during training or through in-context learning—they should not be included in the predicted graph according to our current benchmark criteria. For instance, an LLM might accurately infer a simulation method (such as Density Functional Theory or Molecular Dynamics) from software names mentioned in the text. To mitigate hallucinations, we implemented two primary heuristics. First, we flag as a potential hallucination any value of a datatype property if there is no matching (or even approximately matching) value found in the input text. Second, triples that utilize relations or classes not defined within the input ontology are excluded from the predict graph, even if it is possible to infer them. These strategies prioritize precision over recall for the predicted triples and ensure alignment with the input schema.

    \item[Cleaning and merging of triples and nodes.] This set of heuristics leverages the semantics of the input ontology to generate a coherent knowledge graph. By removing duplicated information and merging partial subgraphs, these techniques are particularly beneficial when extractors in our pipeline process short text fragments, such as sentences. Currently, we employ four specific heuristics. (1) If two nodes share identical property-value pairs, they are merged into a single node. All references to the original nodes in other triples are updated accordingly. (2) If two nodes are instances of the same ontology class, and the property-value pairs of one node are a subset of the property-value pairs of the others, then we remove the former and keep the latter. (3) If two nodes are instances of the same ontology class and contain distinct but non-conflicting information, they are merged into a single node. The node with fewer property-value pairs is integrated into the larger one. (4) For unconnected nodes, we establish connections to other nodes in the graph using defined relations from the ontology, provided this can be done without ambiguity.
\end{description}

The \textit{eolas} pipeline employs consolidation heuristics to transform knowledge graphs derived from individual text fragments into a comprehensive document-level knowledge graph. This transformation process maintains connections to the original sentences or paragraphs where each piece of information was initially extracted.

In our particular use case, various instances of the ontology—such as materials, defect types, measurements, and methodologies—are characterized by multiple properties that are often described across different text segments. Although advanced techniques for merging subgraphs across paragraphs remain an area for future research, our current consolidation heuristics produce a document-level knowledge graph that is seamlessly interconnected at the entity level, and facilitates efficient querying across the entire document (see for example \ref{fig:diagrams-eolas_user_interface_with_graph}(a)).

\subsection{\textit{eolas} user interface}

The user interface offers an intuitive method for selecting a document and exploring its associated knowledge graph (see figure \ref{fig:diagrams-eolas_user_interface_with_graph}(a)). To simplify visualization, the interface presents the graph in a tabular format. In this layout, columns represent the datatype properties of nodes, while rows correspond to instances of these nodes.

Although straightforward, this tabular approach can lead to numerous columns that match all possible datatype properties in the ontology. Consequently, rows often contain many missing values, since each instance typically possesses only a subset of those datatype properties. To address this issue, inclusion/exclusion filters have been implemented. These approach enables a faceted navigation of the tabular data, and allow users to select specific datatype properties they are interested in.

Furthermore, users can choose particular values for each datatype property, effectively narrowing the scope of the tabular visualization and enhancing its utility by focusing on relevant information.

Users can also explore the knowledge graph at a more granular level. They may choose to view subgraphs corresponding to individual text fragments (see figure \ref{fig:diagrams-eolas_user_interface_with_graph}(b)). At this detailed level, the user interface continues to offer a tabular visualization of the knowledge graphs; also, if annotations are available, they are displayed by highlighting relevant sections of the text.

Finally, the user interface supports the visualization of knowledge graphs at both document-level and fragment-level granularity (see figure \ref{fig:diagrams-eolas_user_interface_with_graph}(c)).

\begin{appendices}

\section{Additional results: percentage of success at different thresholds}
\label{sec:appendix:Percentage of success at different thresholds}

We present experimental results using the metric \(P_j(NGED^*)\) (see equation \ref{eq:percentage-of-success}), which represents the percentage of success at various threshold values \(NGED^*\). Specifically, tables \ref{table:ranking-based-on-percentage-of-success-at-nged-0.05}, \ref{table:ranking-based-on-percentage-of-success-at-nged-0.15}, and \ref{table:ranking-based-on-percentage-of-success-at-nged-0.25} report the percentage of success for all 168 configurations (listed in table \ref{table:experiments_summary}) at $NGED^* = 0.05$, $NGED^* = 0.15$, and $NGED^* = 0.25$, respectively.    


\begin{table}
\centering
\caption{Percentage of success 
at normalized graph edit distance $NGED^* = 0.05$ 
for all experimental configurations. Each cell reports the score $P_j(0.05)$ 
(see equation \ref{eq:percentage-of-success}) for the configuration as a percentage.
Colors indicate \colorbox[HTML]{ FFF4B8 }{best}, 
\colorbox[HTML]{ DBDBDB }{2nd best}, and \colorbox[HTML]{ F1DBC6 }{3rd best} 
for each column.}
\label{table:ranking-based-on-percentage-of-success-at-nged-0.05}
\begin{tabular}{rr|rrrr|rrrr}
\toprule
 & & \multicolumn{4}{c|}{zero-shot} & \multicolumn{4}{c}{few-shot} \\
\midrule
 & examples & \multicolumn{4}{c|}{False} & \multicolumn{4}{c}{True} \\
 & instances & \multicolumn{2}{c}{False} & \multicolumn{2}{c|}{True} & \multicolumn{2}{c}{False} & \multicolumn{2}{c}{True} \\
 & consolidation & False & True & False & True & False & True & False & True \\
\makecell[r]{instructions\\serialization} & LLM &  &  &  &  &  &  &  &  \\
\midrule
\multirow[c]{6}{*}{S, TTL} & GPT\_120 & {\cellcolor[HTML]{FFF4B8}} 12.6 & {\cellcolor[HTML]{FFF4B8}} 12.6 & {\cellcolor[HTML]{FFF4B8}} 12.6 & {\cellcolor[HTML]{FFF4B8}} 12.6 & {\cellcolor[HTML]{F1DBC6}} 27.9 & {\cellcolor[HTML]{F1DBC6}} 28.8 & 19.8 & 19.8 \\
 & GPT\_20 & {\cellcolor[HTML]{FFF4B8}} 12.6 & {\cellcolor[HTML]{FFF4B8}} 12.6 & {\cellcolor[HTML]{FFF4B8}} 12.6 & {\cellcolor[HTML]{FFF4B8}} 12.6 & 5.4 & 5.4 & 0.9 & 1.8 \\
 & GRANITE & 4.5 & 5.4 & 1.8 & 2.7 & 7.2 & 8.1 & 9.0 & 9.9 \\
 & LLAMA\_3 & {\cellcolor[HTML]{DBDBDB}} 11.7 & {\cellcolor[HTML]{DBDBDB}} 11.7 & 7.2 & {\cellcolor[HTML]{F1DBC6}} 9.9 & 20.7 & 22.5 & 24.3 & 26.1 \\
 & LLAMA\_4 & {\cellcolor[HTML]{DBDBDB}} 11.7 & {\cellcolor[HTML]{DBDBDB}} 11.7 & {\cellcolor[HTML]{DBDBDB}} 9.9 & {\cellcolor[HTML]{FFF4B8}} 12.6 & {\cellcolor[HTML]{DBDBDB}} 28.8 & {\cellcolor[HTML]{F1DBC6}} 28.8 & 19.8 & 19.8 \\
 & MISTRAL & 2.7 & 2.7 & 6.3 & {\cellcolor[HTML]{DBDBDB}} 10.8 & 21.6 & 24.3 & 22.5 & 24.3 \\
\cmidrule{1-10}
\multirow[c]{6}{*}{D, TTL} & GPT\_120 & {\cellcolor[HTML]{FFF4B8}} 12.6 & {\cellcolor[HTML]{FFF4B8}} 12.6 & {\cellcolor[HTML]{FFF4B8}} 12.6 & {\cellcolor[HTML]{FFF4B8}} 12.6 & {\cellcolor[HTML]{F1DBC6}} 27.9 & {\cellcolor[HTML]{F1DBC6}} 28.8 & 21.6 & 23.4 \\
 & GPT\_20 & {\cellcolor[HTML]{FFF4B8}} 12.6 & {\cellcolor[HTML]{FFF4B8}} 12.6 & {\cellcolor[HTML]{FFF4B8}} 12.6 & {\cellcolor[HTML]{FFF4B8}} 12.6 & 19.8 & 21.6 & 15.3 & 15.3 \\
 & GRANITE & 7.2 & 8.1 & 0.9 & 0.9 & 5.4 & 7.2 & 9.9 & 9.9 \\
 & LLAMA\_3 & 6.3 & 9.0 & 6.3 & {\cellcolor[HTML]{DBDBDB}} 10.8 & 25.2 & {\cellcolor[HTML]{F1DBC6}} 28.8 & {\cellcolor[HTML]{FFF4B8}} 27.9 & {\cellcolor[HTML]{FFF4B8}} 33.3 \\
 & LLAMA\_4 & {\cellcolor[HTML]{F1DBC6}} 10.8 & {\cellcolor[HTML]{F1DBC6}} 10.8 & {\cellcolor[HTML]{DBDBDB}} 9.9 & {\cellcolor[HTML]{F1DBC6}} 9.9 & 25.2 & 26.1 & 23.4 & 23.4 \\
 & MISTRAL & 5.4 & 6.3 & 4.5 & 5.4 & {\cellcolor[HTML]{FFF4B8}} 29.7 & {\cellcolor[HTML]{FFF4B8}} 34.2 & {\cellcolor[HTML]{DBDBDB}} 27.0 & {\cellcolor[HTML]{DBDBDB}} 32.4 \\
\cmidrule{1-10}
\multirow[c]{6}{*}{S, BAML} & GPT\_120 & 4.5 & 5.4 & 4.5 & 4.5 & {\cellcolor[HTML]{DBDBDB}} 28.8 & {\cellcolor[HTML]{DBDBDB}} 29.7 & {\cellcolor[HTML]{F1DBC6}} 26.1 & {\cellcolor[HTML]{F1DBC6}} 27.9 \\
 & GPT\_20 & 4.5 & 4.5 & {\cellcolor[HTML]{F1DBC6}} 8.1 & 9.0 & 18.0 & 18.9 & 19.8 & 20.7 \\
 & GRANITE & 0.9 & 3.6 & 2.7 & 5.4 & 15.3 & 16.2 & 16.2 & 21.6 \\
 & LLAMA\_3 & 2.7 & 4.5 & 3.6 & 6.3 & 18.0 & 19.8 & {\cellcolor[HTML]{F1DBC6}} 26.1 & {\cellcolor[HTML]{F1DBC6}} 27.9 \\
 & LLAMA\_4 & 1.8 & 5.4 & 4.5 & 9.0 & 15.3 & 18.9 & 11.7 & 16.2 \\
 & MISTRAL & 4.5 & 4.5 & 5.4 & 5.4 & {\cellcolor[HTML]{F1DBC6}} 27.9 & {\cellcolor[HTML]{F1DBC6}} 28.8 & 25.2 & 27.0 \\
\cmidrule{1-10}
\multirow[c]{6}{*}{S, VERB} & GPT\_120 & \color[HTML]{FFFFFF} nan & \color[HTML]{FFFFFF} nan & \color[HTML]{FFFFFF} nan & \color[HTML]{FFFFFF} nan & 27.0 & {\cellcolor[HTML]{F1DBC6}} 28.8 & \color[HTML]{FFFFFF} nan & \color[HTML]{FFFFFF} nan \\
 & GPT\_20 & \color[HTML]{FFFFFF} nan & \color[HTML]{FFFFFF} nan & \color[HTML]{FFFFFF} nan & \color[HTML]{FFFFFF} nan & 5.4 & 5.4 & \color[HTML]{FFFFFF} nan & \color[HTML]{FFFFFF} nan \\
 & GRANITE & \color[HTML]{FFFFFF} nan & \color[HTML]{FFFFFF} nan & \color[HTML]{FFFFFF} nan & \color[HTML]{FFFFFF} nan & 5.4 & 5.4 & \color[HTML]{FFFFFF} nan & \color[HTML]{FFFFFF} nan \\
 & LLAMA\_3 & \color[HTML]{FFFFFF} nan & \color[HTML]{FFFFFF} nan & \color[HTML]{FFFFFF} nan & \color[HTML]{FFFFFF} nan & 17.1 & 18.9 & \color[HTML]{FFFFFF} nan & \color[HTML]{FFFFFF} nan \\
 & LLAMA\_4 & \color[HTML]{FFFFFF} nan & \color[HTML]{FFFFFF} nan & \color[HTML]{FFFFFF} nan & \color[HTML]{FFFFFF} nan & 22.5 & 23.4 & \color[HTML]{FFFFFF} nan & \color[HTML]{FFFFFF} nan \\
 & MISTRAL & \color[HTML]{FFFFFF} nan & \color[HTML]{FFFFFF} nan & \color[HTML]{FFFFFF} nan & \color[HTML]{FFFFFF} nan & 18.9 & 20.7 & \color[HTML]{FFFFFF} nan & \color[HTML]{FFFFFF} nan \\
\cmidrule{1-10}
\multirow[c]{6}{*}{D, VERB} & GPT\_120 & \color[HTML]{FFFFFF} nan & \color[HTML]{FFFFFF} nan & \color[HTML]{FFFFFF} nan & \color[HTML]{FFFFFF} nan & 25.2 & 25.2 & \color[HTML]{FFFFFF} nan & \color[HTML]{FFFFFF} nan \\
 & GPT\_20 & \color[HTML]{FFFFFF} nan & \color[HTML]{FFFFFF} nan & \color[HTML]{FFFFFF} nan & \color[HTML]{FFFFFF} nan & 17.1 & 18.9 & \color[HTML]{FFFFFF} nan & \color[HTML]{FFFFFF} nan \\
 & GRANITE & \color[HTML]{FFFFFF} nan & \color[HTML]{FFFFFF} nan & \color[HTML]{FFFFFF} nan & \color[HTML]{FFFFFF} nan & 6.3 & 7.2 & \color[HTML]{FFFFFF} nan & \color[HTML]{FFFFFF} nan \\
 & LLAMA\_3 & \color[HTML]{FFFFFF} nan & \color[HTML]{FFFFFF} nan & \color[HTML]{FFFFFF} nan & \color[HTML]{FFFFFF} nan & 20.7 & 21.6 & \color[HTML]{FFFFFF} nan & \color[HTML]{FFFFFF} nan \\
 & LLAMA\_4 & \color[HTML]{FFFFFF} nan & \color[HTML]{FFFFFF} nan & \color[HTML]{FFFFFF} nan & \color[HTML]{FFFFFF} nan & 24.3 & 24.3 & \color[HTML]{FFFFFF} nan & \color[HTML]{FFFFFF} nan \\
 & MISTRAL & \color[HTML]{FFFFFF} nan & \color[HTML]{FFFFFF} nan & \color[HTML]{FFFFFF} nan & \color[HTML]{FFFFFF} nan & 27.0 & {\cellcolor[HTML]{F1DBC6}} 28.8 & \color[HTML]{FFFFFF} nan & \color[HTML]{FFFFFF} nan \\
\cmidrule{1-10}
\bottomrule
\end{tabular}
\end{table}


\begin{table}
\centering
\caption{Percentage of success 
at normalized graph edit distance $NGED^* = 0.15$ 
for all experimental configurations. Each cell reports the score $P_j(0.15)$ 
(see equation \ref{eq:percentage-of-success}) for the configuration as a percentage.
Colors indicate \colorbox[HTML]{ FFF4B8 }{best}, 
\colorbox[HTML]{ DBDBDB }{2nd best}, and \colorbox[HTML]{ F1DBC6 }{3rd best} 
for each column.}
\label{table:ranking-based-on-percentage-of-success-at-nged-0.15}
\begin{tabular}{rr|rrrr|rrrr}
\toprule
 & & \multicolumn{4}{c|}{zero-shot} & \multicolumn{4}{c}{few-shot} \\
\midrule
 & examples & \multicolumn{4}{c|}{False} & \multicolumn{4}{c}{True} \\
 & instances & \multicolumn{2}{c}{False} & \multicolumn{2}{c|}{True} & \multicolumn{2}{c}{False} & \multicolumn{2}{c}{True} \\
 & consolidation & False & True & False & True & False & True & False & True \\
\makecell[r]{instructions\\serialization} & LLM &  &  &  &  &  &  &  &  \\
\midrule
\multirow[c]{6}{*}{S, TTL} & GPT\_120 & {\cellcolor[HTML]{F1DBC6}} 12.6 & 12.6 & 12.6 & 12.6 & {\cellcolor[HTML]{DBDBDB}} 53.2 & 52.3 & {\cellcolor[HTML]{F1DBC6}} 48.6 & 50.5 \\
 & GPT\_20 & {\cellcolor[HTML]{F1DBC6}} 12.6 & 12.6 & 12.6 & 12.6 & 19.8 & 21.6 & 16.2 & 18.9 \\
 & GRANITE & 4.5 & 5.4 & 1.8 & 2.7 & 18.9 & 19.8 & 18.9 & 19.8 \\
 & LLAMA\_3 & 11.7 & 13.5 & 7.2 & 9.9 & 36.0 & 38.7 & 39.6 & 42.3 \\
 & LLAMA\_4 & {\cellcolor[HTML]{F1DBC6}} 12.6 & 14.4 & 15.3 & 18.0 & 39.6 & 39.6 & 31.5 & 32.4 \\
 & MISTRAL & 4.5 & 4.5 & 9.9 & 16.2 & 40.5 & 42.3 & 41.4 & 45.9 \\
\cmidrule{1-10}
\multirow[c]{6}{*}{D, TTL} & GPT\_120 & {\cellcolor[HTML]{F1DBC6}} 12.6 & 12.6 & 12.6 & 12.6 & 50.5 & {\cellcolor[HTML]{F1DBC6}} 53.2 & 46.8 & 48.6 \\
 & GPT\_20 & {\cellcolor[HTML]{F1DBC6}} 12.6 & 12.6 & 12.6 & 12.6 & 43.2 & 47.7 & 42.3 & 42.3 \\
 & GRANITE & 7.2 & 8.1 & 0.9 & 0.9 & 14.4 & 16.2 & 20.7 & 20.7 \\
 & LLAMA\_3 & 6.3 & 9.0 & 9.9 & 15.3 & 42.3 & 45.0 & 47.7 & {\cellcolor[HTML]{F1DBC6}} 53.2 \\
 & LLAMA\_4 & 11.7 & 15.3 & 18.0 & 18.9 & 43.2 & 45.0 & 39.6 & 37.8 \\
 & MISTRAL & 6.3 & 9.0 & {\cellcolor[HTML]{DBDBDB}} 25.2 & {\cellcolor[HTML]{DBDBDB}} 26.1 & 47.7 & 50.5 & {\cellcolor[HTML]{F1DBC6}} 48.6 & 52.3 \\
\cmidrule{1-10}
\multirow[c]{6}{*}{S, BAML} & GPT\_120 & {\cellcolor[HTML]{FFF4B8}} 22.5 & {\cellcolor[HTML]{FFF4B8}} 25.2 & {\cellcolor[HTML]{FFF4B8}} 26.1 & {\cellcolor[HTML]{FFF4B8}} 27.0 & 47.7 & 48.6 & {\cellcolor[HTML]{FFF4B8}} 55.0 & {\cellcolor[HTML]{FFF4B8}} 58.6 \\
 & GPT\_20 & {\cellcolor[HTML]{DBDBDB}} 15.3 & {\cellcolor[HTML]{F1DBC6}} 16.2 & {\cellcolor[HTML]{F1DBC6}} 20.7 & {\cellcolor[HTML]{F1DBC6}} 23.4 & 47.7 & 47.7 & 45.9 & 46.8 \\
 & GRANITE & 2.7 & 7.2 & 4.5 & 8.1 & 26.1 & 27.9 & 29.7 & 36.0 \\
 & LLAMA\_3 & {\cellcolor[HTML]{F1DBC6}} 12.6 & {\cellcolor[HTML]{DBDBDB}} 17.1 & 12.6 & 17.1 & 42.3 & 45.0 & {\cellcolor[HTML]{DBDBDB}} 51.4 & {\cellcolor[HTML]{DBDBDB}} 55.9 \\
 & LLAMA\_4 & 4.5 & 14.4 & 8.1 & {\cellcolor[HTML]{DBDBDB}} 26.1 & 36.9 & 42.3 & 17.1 & 24.3 \\
 & MISTRAL & 8.1 & 9.0 & 11.7 & 12.6 & 47.7 & 52.3 & {\cellcolor[HTML]{F1DBC6}} 48.6 & {\cellcolor[HTML]{F1DBC6}} 53.2 \\
\cmidrule{1-10}
\multirow[c]{6}{*}{S, VERB} & GPT\_120 & \color[HTML]{FFFFFF} nan & \color[HTML]{FFFFFF} nan & \color[HTML]{FFFFFF} nan & \color[HTML]{FFFFFF} nan & {\cellcolor[HTML]{FFF4B8}} 55.9 & {\cellcolor[HTML]{FFF4B8}} 56.8 & \color[HTML]{FFFFFF} nan & \color[HTML]{FFFFFF} nan \\
 & GPT\_20 & \color[HTML]{FFFFFF} nan & \color[HTML]{FFFFFF} nan & \color[HTML]{FFFFFF} nan & \color[HTML]{FFFFFF} nan & 16.2 & 18.0 & \color[HTML]{FFFFFF} nan & \color[HTML]{FFFFFF} nan \\
 & GRANITE & \color[HTML]{FFFFFF} nan & \color[HTML]{FFFFFF} nan & \color[HTML]{FFFFFF} nan & \color[HTML]{FFFFFF} nan & 9.0 & 11.7 & \color[HTML]{FFFFFF} nan & \color[HTML]{FFFFFF} nan \\
 & LLAMA\_3 & \color[HTML]{FFFFFF} nan & \color[HTML]{FFFFFF} nan & \color[HTML]{FFFFFF} nan & \color[HTML]{FFFFFF} nan & 34.2 & 38.7 & \color[HTML]{FFFFFF} nan & \color[HTML]{FFFFFF} nan \\
 & LLAMA\_4 & \color[HTML]{FFFFFF} nan & \color[HTML]{FFFFFF} nan & \color[HTML]{FFFFFF} nan & \color[HTML]{FFFFFF} nan & 37.8 & 39.6 & \color[HTML]{FFFFFF} nan & \color[HTML]{FFFFFF} nan \\
 & MISTRAL & \color[HTML]{FFFFFF} nan & \color[HTML]{FFFFFF} nan & \color[HTML]{FFFFFF} nan & \color[HTML]{FFFFFF} nan & 38.7 & 41.4 & \color[HTML]{FFFFFF} nan & \color[HTML]{FFFFFF} nan \\
\cmidrule{1-10}
\multirow[c]{6}{*}{D, VERB} & GPT\_120 & \color[HTML]{FFFFFF} nan & \color[HTML]{FFFFFF} nan & \color[HTML]{FFFFFF} nan & \color[HTML]{FFFFFF} nan & {\cellcolor[HTML]{F1DBC6}} 51.4 & {\cellcolor[HTML]{DBDBDB}} 54.1 & \color[HTML]{FFFFFF} nan & \color[HTML]{FFFFFF} nan \\
 & GPT\_20 & \color[HTML]{FFFFFF} nan & \color[HTML]{FFFFFF} nan & \color[HTML]{FFFFFF} nan & \color[HTML]{FFFFFF} nan & 37.8 & 42.3 & \color[HTML]{FFFFFF} nan & \color[HTML]{FFFFFF} nan \\
 & GRANITE & \color[HTML]{FFFFFF} nan & \color[HTML]{FFFFFF} nan & \color[HTML]{FFFFFF} nan & \color[HTML]{FFFFFF} nan & 16.2 & 18.0 & \color[HTML]{FFFFFF} nan & \color[HTML]{FFFFFF} nan \\
 & LLAMA\_3 & \color[HTML]{FFFFFF} nan & \color[HTML]{FFFFFF} nan & \color[HTML]{FFFFFF} nan & \color[HTML]{FFFFFF} nan & 41.4 & 42.3 & \color[HTML]{FFFFFF} nan & \color[HTML]{FFFFFF} nan \\
 & LLAMA\_4 & \color[HTML]{FFFFFF} nan & \color[HTML]{FFFFFF} nan & \color[HTML]{FFFFFF} nan & \color[HTML]{FFFFFF} nan & 39.6 & 41.4 & \color[HTML]{FFFFFF} nan & \color[HTML]{FFFFFF} nan \\
 & MISTRAL & \color[HTML]{FFFFFF} nan & \color[HTML]{FFFFFF} nan & \color[HTML]{FFFFFF} nan & \color[HTML]{FFFFFF} nan & 49.5 & {\cellcolor[HTML]{F1DBC6}} 53.2 & \color[HTML]{FFFFFF} nan & \color[HTML]{FFFFFF} nan \\
\cmidrule{1-10}
\bottomrule
\end{tabular}
\end{table}


\begin{table}
\centering
\caption{Percentage of success 
at normalized graph edit distance $NGED^* = 0.25$ 
for all experimental configurations. Each cell reports the score $P_j(0.25)$ 
(see equation \ref{eq:percentage-of-success}) for the configuration as a percentage.
Colors indicate \colorbox[HTML]{ FFF4B8 }{best}, 
\colorbox[HTML]{ DBDBDB }{2nd best}, and \colorbox[HTML]{ F1DBC6 }{3rd best} 
for each column.}
\label{table:ranking-based-on-percentage-of-success-at-nged-0.25}
\begin{tabular}{rr|rrrr|rrrr}
\toprule
 & & \multicolumn{4}{c|}{zero-shot} & \multicolumn{4}{c}{few-shot} \\
\midrule
 & examples & \multicolumn{4}{c|}{False} & \multicolumn{4}{c}{True} \\
 & instances & \multicolumn{2}{c}{False} & \multicolumn{2}{c|}{True} & \multicolumn{2}{c}{False} & \multicolumn{2}{c}{True} \\
 & consolidation & False & True & False & True & False & True & False & True \\
\makecell[r]{instructions\\serialization} & LLM &  &  &  &  &  &  &  &  \\
\midrule
\multirow[c]{6}{*}{S, TTL} & GPT\_120 & 12.6 & 12.6 & 12.6 & 12.6 & {\cellcolor[HTML]{F1DBC6}} 67.6 & {\cellcolor[HTML]{DBDBDB}} 69.4 & 65.8 & {\cellcolor[HTML]{F1DBC6}} 67.6 \\
 & GPT\_20 & 12.6 & 12.6 & 12.6 & 12.6 & 36.9 & 36.9 & 37.8 & 38.7 \\
 & GRANITE & 7.2 & 8.1 & 2.7 & 3.6 & 31.5 & 33.3 & 29.7 & 31.5 \\
 & LLAMA\_3 & 12.6 & 14.4 & 9.0 & 11.7 & 49.5 & 51.4 & 55.9 & 57.7 \\
 & LLAMA\_4 & 16.2 & 19.8 & 27.9 & 31.5 & 48.6 & 48.6 & 44.1 & 45.0 \\
 & MISTRAL & 8.1 & 16.2 & 16.2 & 24.3 & 55.0 & 54.1 & 56.8 & 56.8 \\
\cmidrule{1-10}
\multirow[c]{6}{*}{D, TTL} & GPT\_120 & 12.6 & 12.6 & 12.6 & 12.6 & {\cellcolor[HTML]{DBDBDB}} 69.4 & {\cellcolor[HTML]{FFF4B8}} 70.3 & 62.2 & 65.8 \\
 & GPT\_20 & 12.6 & 12.6 & 12.6 & 12.6 & 57.7 & 62.2 & 59.5 & 64.0 \\
 & GRANITE & 8.1 & 9.0 & 0.9 & 0.9 & 29.7 & 35.1 & 34.2 & 36.9 \\
 & LLAMA\_3 & 10.8 & 13.5 & 15.3 & 22.5 & 60.4 & 59.5 & {\cellcolor[HTML]{F1DBC6}} 67.6 & {\cellcolor[HTML]{DBDBDB}} 70.3 \\
 & LLAMA\_4 & 12.6 & {\cellcolor[HTML]{F1DBC6}} 26.1 & 30.6 & 30.6 & 57.7 & 59.5 & 52.3 & 51.4 \\
 & MISTRAL & 10.8 & 19.8 & {\cellcolor[HTML]{DBDBDB}} 41.4 & {\cellcolor[HTML]{DBDBDB}} 42.3 & {\cellcolor[HTML]{F1DBC6}} 67.6 & 67.6 & 64.0 & 66.7 \\
\cmidrule{1-10}
\multirow[c]{6}{*}{S, BAML} & GPT\_120 & {\cellcolor[HTML]{DBDBDB}} 30.6 & {\cellcolor[HTML]{FFF4B8}} 34.2 & {\cellcolor[HTML]{FFF4B8}} 45.0 & {\cellcolor[HTML]{FFF4B8}} 44.1 & 64.9 & 63.1 & {\cellcolor[HTML]{DBDBDB}} 69.4 & {\cellcolor[HTML]{DBDBDB}} 70.3 \\
 & GPT\_20 & {\cellcolor[HTML]{FFF4B8}} 31.5 & {\cellcolor[HTML]{FFF4B8}} 34.2 & {\cellcolor[HTML]{F1DBC6}} 40.5 & {\cellcolor[HTML]{F1DBC6}} 39.6 & 60.4 & 61.3 & {\cellcolor[HTML]{FFF4B8}} 72.1 & {\cellcolor[HTML]{FFF4B8}} 73.0 \\
 & GRANITE & 9.9 & 14.4 & 8.1 & 12.6 & 38.7 & 39.6 & 44.1 & 54.1 \\
 & LLAMA\_3 & {\cellcolor[HTML]{F1DBC6}} 21.6 & {\cellcolor[HTML]{DBDBDB}} 27.9 & 22.5 & 27.9 & 61.3 & 62.2 & {\cellcolor[HTML]{F1DBC6}} 67.6 & {\cellcolor[HTML]{F1DBC6}} 67.6 \\
 & LLAMA\_4 & 13.5 & 23.4 & 21.6 & 35.1 & 54.1 & 57.7 & 23.4 & 27.0 \\
 & MISTRAL & 15.3 & 19.8 & 18.9 & 17.1 & 64.9 & 66.7 & 63.1 & 64.9 \\
\cmidrule{1-10}
\multirow[c]{6}{*}{S, VERB} & GPT\_120 & \color[HTML]{FFFFFF} nan & \color[HTML]{FFFFFF} nan & \color[HTML]{FFFFFF} nan & \color[HTML]{FFFFFF} nan & {\cellcolor[HTML]{FFF4B8}} 72.1 & {\cellcolor[HTML]{FFF4B8}} 70.3 & \color[HTML]{FFFFFF} nan & \color[HTML]{FFFFFF} nan \\
 & GPT\_20 & \color[HTML]{FFFFFF} nan & \color[HTML]{FFFFFF} nan & \color[HTML]{FFFFFF} nan & \color[HTML]{FFFFFF} nan & 27.9 & 32.4 & \color[HTML]{FFFFFF} nan & \color[HTML]{FFFFFF} nan \\
 & GRANITE & \color[HTML]{FFFFFF} nan & \color[HTML]{FFFFFF} nan & \color[HTML]{FFFFFF} nan & \color[HTML]{FFFFFF} nan & 23.4 & 25.2 & \color[HTML]{FFFFFF} nan & \color[HTML]{FFFFFF} nan \\
 & LLAMA\_3 & \color[HTML]{FFFFFF} nan & \color[HTML]{FFFFFF} nan & \color[HTML]{FFFFFF} nan & \color[HTML]{FFFFFF} nan & 53.2 & 55.9 & \color[HTML]{FFFFFF} nan & \color[HTML]{FFFFFF} nan \\
 & LLAMA\_4 & \color[HTML]{FFFFFF} nan & \color[HTML]{FFFFFF} nan & \color[HTML]{FFFFFF} nan & \color[HTML]{FFFFFF} nan & 47.7 & 49.5 & \color[HTML]{FFFFFF} nan & \color[HTML]{FFFFFF} nan \\
 & MISTRAL & \color[HTML]{FFFFFF} nan & \color[HTML]{FFFFFF} nan & \color[HTML]{FFFFFF} nan & \color[HTML]{FFFFFF} nan & 55.0 & 57.7 & \color[HTML]{FFFFFF} nan & \color[HTML]{FFFFFF} nan \\
\cmidrule{1-10}
\multirow[c]{6}{*}{D, VERB} & GPT\_120 & \color[HTML]{FFFFFF} nan & \color[HTML]{FFFFFF} nan & \color[HTML]{FFFFFF} nan & \color[HTML]{FFFFFF} nan & 66.7 & {\cellcolor[HTML]{F1DBC6}} 68.5 & \color[HTML]{FFFFFF} nan & \color[HTML]{FFFFFF} nan \\
 & GPT\_20 & \color[HTML]{FFFFFF} nan & \color[HTML]{FFFFFF} nan & \color[HTML]{FFFFFF} nan & \color[HTML]{FFFFFF} nan & 54.1 & 57.7 & \color[HTML]{FFFFFF} nan & \color[HTML]{FFFFFF} nan \\
 & GRANITE & \color[HTML]{FFFFFF} nan & \color[HTML]{FFFFFF} nan & \color[HTML]{FFFFFF} nan & \color[HTML]{FFFFFF} nan & 28.8 & 34.2 & \color[HTML]{FFFFFF} nan & \color[HTML]{FFFFFF} nan \\
 & LLAMA\_3 & \color[HTML]{FFFFFF} nan & \color[HTML]{FFFFFF} nan & \color[HTML]{FFFFFF} nan & \color[HTML]{FFFFFF} nan & 56.8 & 57.7 & \color[HTML]{FFFFFF} nan & \color[HTML]{FFFFFF} nan \\
 & LLAMA\_4 & \color[HTML]{FFFFFF} nan & \color[HTML]{FFFFFF} nan & \color[HTML]{FFFFFF} nan & \color[HTML]{FFFFFF} nan & 51.4 & 51.4 & \color[HTML]{FFFFFF} nan & \color[HTML]{FFFFFF} nan \\
 & MISTRAL & \color[HTML]{FFFFFF} nan & \color[HTML]{FFFFFF} nan & \color[HTML]{FFFFFF} nan & \color[HTML]{FFFFFF} nan & {\cellcolor[HTML]{F1DBC6}} 67.6 & {\cellcolor[HTML]{DBDBDB}} 69.4 & \color[HTML]{FFFFFF} nan & \color[HTML]{FFFFFF} nan \\
\cmidrule{1-10}
\bottomrule
\end{tabular}
\end{table}

\clearpage

\section{Competency Questions and Graph-RAG}\label{secA2}

Competency questions (CQs) are a standard mechanism for evaluating whether an ontology can represent the knowledge required for its intended purpose. By assessing whether a knowledge graph (KG) constructed using the ontology can answer these questions, one can determine whether the ontology captures the necessary domain concepts, relationships, and contextual information.

Ontologies —and the KGs built from them— play a critical role in addressing challenges posed by the heterogeneity and inconsistency of scientific reporting. By normalizing terminology and encoding semantic relationships, they support cross-document queries and enable logical inference. Axioms embedded within the ontology can help infer missing data, enforce domain constraints, and identify implausible or hallucinated outputs. Because KGs adhere to a shared schema, they can be seamlessly integrated into advanced downstream applications such as question answering, specifically multi-hop, graph-based retrieval-augmented generation, or Graph-RAG, and support evidence-backed model interpretation.

In our work, the ontology was refined primarily to capture the contextual information required during ground-truth construction. The resulting document-level KGs can be rendered as tabular data, queried uniformly using SPARQL, or incorporated into existing Graph-RAG pipelines \cite{edge2024from, wang2024graph}. 

\begin{figure}[ht]
    \centering
    \includegraphics[width=\linewidth]{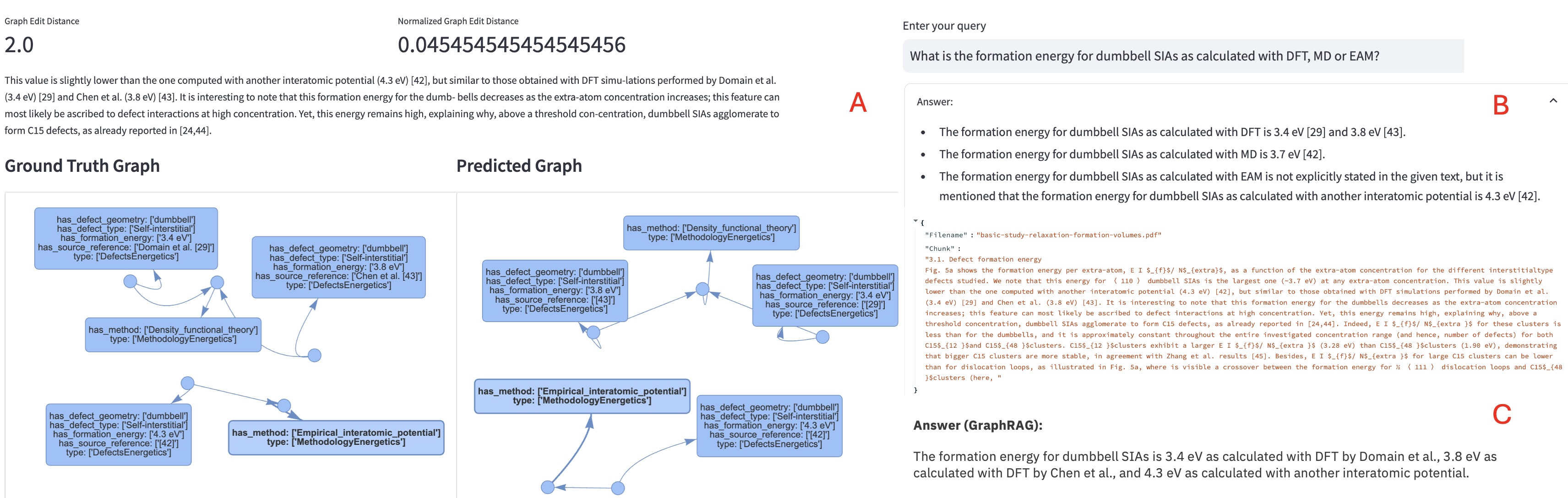}
    \caption{(A) Ground truth and predicted graphs generated by LLaMA-3 in few-shot (using schema-only prompts with a GED = 2.0 in between both graphs). Multiple formation energy values are associated with distinct methods and reference sources, preserving fidelity to the original text structure. A standard RAG approached (B) failed to answer the question "What is the formation energy for dumbbell SIAs as calculated with DFT, MD or EAM?", despite retrieving the correct document and passage. A naive Graph-RAG approach (C), which prompts an LLM to retrieve an answer, using as input either the full ground truth KG or the document-level predicted one containing that passage, retrieves the correct answer.}  
    \label{fig:example-rag}
\end{figure}

Figure \ref{fig:example-rag} provides an example of both a ground truth KG and a predicted KG extracted from the same text passage. These KGs contains all the information needed to successfully answer a domain-specific multi-hop question using a naive Graph-RAG approach in which an LLM is prompted with the KG rather than raw text.

A set of expert-defined CQs and their corresponding answers —generated from the 126 manually curated passages used to build the ground truth— are provided in the following table \ref{tab:competency-questions}, illustrating the value of structured knowledge in answering complex queries that span multiple paragraphs or documents.

These questions were answered using a simple Graph-RAG method applied over the ground-truth KGs, including examples used during few-shot prompting. The complete ground-truth KG from the 126 passages is small enough to fit within the context window of a large LLM, eliminating the need for a retrieval step to construct a subgraph prior to answering a question. This simple evaluation demonstrates that the ontology-compliant knowledge graphs, and, by extension, the ontology itself, can successfully support complex information needs that span multiple paragraphs or documents. The examples highlight the value of structured knowledge in enabling reliable, interpretable querying of scientific literature given a user defined-schema, and illustrate their effectiveness to provide accurate answers.

{\small
\begin{longtable}
{|p{0.45\textwidth}|p{0.5\textwidth}|}
\caption{Some example competency questions and their answers using Graph-RAG. NA is retrieved correctly when the answer is not in the input KG.}
\label{tab:competency-questions} \\
\hline
\textbf{Question} & \textbf{Answer (Graph-RAG)} \\
\hline
\endfirsthead

\hline
\textbf{Question} & \textbf{Answer (Graph-RAG)} \\
\hline
\endhead

\hline
\endfoot

\hline
\endlastfoot

What is the relaxation volume of C15 clusters of size 12 and 48? &
The relaxation volume of C15 clusters of size 12 and 48 calculated with DFT is
0.91 $\Omega$ and 0.98 $\Omega$, respectively. \\
\hline

What is the relaxation volume for all defect types? &
The relaxation volume for all defect types is as follows:
\begin{tabular}[t]{@{}l@{}}
-- Dumbbell SIA: 1.5 $\Omega$ \\
-- C15 clusters (size 12): 0.91 $\Omega$ \\
-- C15 clusters (size 48): 0.98 $\Omega$ \\
-- $\frac{1}{2}\langle 111\rangle$ dislocation loops: 0.905 $\Omega$ \\
-- $\langle 100\rangle$ dislocation loops: 0.873 $\Omega$ \\
-- Single vacancies: $-0.11$ $\Omega$ \\
-- Spherical cavities (radius 1 nm): 0.46 eV \\
-- L100 and L111 defects: 0.95 $\Omega$ \\
-- Loop 100 (size 6): 0.86 $\Omega$ \\
-- Loop 111 (size 6): 0.89 $\Omega$ \\
\ldots\ etc.
\end{tabular} \\
\hline

What is the relaxation volume for all defect types as calculated with DFT? &
The relaxation volume for all defect types as calculated with DFT is not explicitly
provided in the knowledge graph. However, the relaxation volume for specific defect
types such as dumbbell SIAs and C15 clusters is mentioned. For dumbbell SIAs, the
relaxation volume is 1.62 $\Omega_0$ according to DFT simulations by Ma et al. For
C15 clusters, the relaxation volume is 0.91 $\Omega_0$ for C15$_{12}$ and 0.98
$\Omega_0$ for C15$_{48}$. \\
\hline

What is the migration energy of a vacancy in iron according to DFT calculations? &
The migration energy of a vacancy in iron according to DFT calculations is 0.65 eV. \\
\hline

What is the migration energy of a self-interstitial in iron according to DFT calculations? &
The migration energy of a self-interstitial in iron according to DFT calculations is
0.3 eV. \\
\hline

What is the migration energy of a vacancy in iron according to molecular dynamics using empirical potentials? &
NA \\
\hline

What is the migration energy of a self-interstitial in iron according to molecular dynamics using empirical potentials? &
The migration energy of a self-interstitial in iron according to molecular dynamics
using empirical potentials is 0.31 eV. \\
\hline

What are the values of the migration energy of vacancy clusters in iron of different sizes? &
The migration energy of vacancy clusters in iron varies with the size of the cluster.
Here are some specific values: V2: 0.62 eV, V3: 0.35 eV, V4: 0.48 eV, V50: $\sim$1.5 eV,
V90: $\sim$1 eV, $>$5 vacancies: 0.9--1.6 eV. \\
\hline

What are the values of the migration energy of interstitial clusters in iron of different sizes? &
The migration energy of interstitial clusters in iron varies depending on the size
and configuration of the clusters. Here are some specific values: I2: 0.42 $\pm$ 0.03 eV,
I3: $\sim$0.43 eV, $<$5 dumbbells: 0.3--0.4 eV, larger clusters: up to $\sim$1.5 eV. \\
\hline

How many different types of defects can be found in irradiated iron? &
NA \\
\hline

What is the migration energy of $\frac{1}{2}\langle 111\rangle$ clusters in iron? &
The migration energy of $\frac{1}{2}\langle 111\rangle$ clusters in iron is approximately
2.0 eV. \\
\hline

What is the migration energy of $\langle 100\rangle$ clusters in iron? &
Approximately 2.0 eV. \\
\hline

What are the values of the binding energies of vacancy clusters in iron of different sizes? &
The binding energies of vacancy clusters in iron of different sizes are as follows:
V2: 0.30 eV, V3: 0.37 eV, V4: 0.62 eV, $>$5 vacancies: 0.9--1.6 eV. \\
\hline

What are the values of the binding energies of interstitial clusters in iron of different sizes? &
The binding energies of interstitial clusters in iron of different sizes are as follows:
I2: 0.80 eV, I3: 0.92 eV, I4: 1.64 eV. \\
\hline

What are the values of the formation energies of vacancy clusters in iron of different sizes? &
The formation energies of vacancy clusters in iron of different sizes are as follows:
102-vacancy void: 58.1 eV, 97-vacancy loop: 114 eV (elastic), 111.1 eV (atomistic),
V90: $\sim$1 eV, V50: $\sim$1.5 eV, V4: 0.48 eV, V3: 0.35 eV, V2: 0.62 eV, V1: 0.67 eV. \\
\hline

\end{longtable}
}

\clearpage

\section{Effect of prompting strategies}\label{secA3}

Given an ontology and a configuration, \textit{eolas} automatically generates prompts for extracting structured knowledge from text passages, either from the documents or from the Ground Truth for evaluation purposes, and independent of the domain or ontology used.  
All prompts consist of: (i) a configurable instruction, (ii) a representation of the ontology, (iii) few-shot examples derived from the KG demonstrating an input passage and the corresponding triples (serialized in turtle); and (iv) the target passage to be processed, prefixed with ”Context:”. The prompt ends with a ”Triples:”, where the model is expected to generate the output following the same convention as in the given examples. Note that "Context:" is also given as the stop sequence for all models.

\begin{figure}
\centering
\noindent\includegraphics[width=1\textwidth]{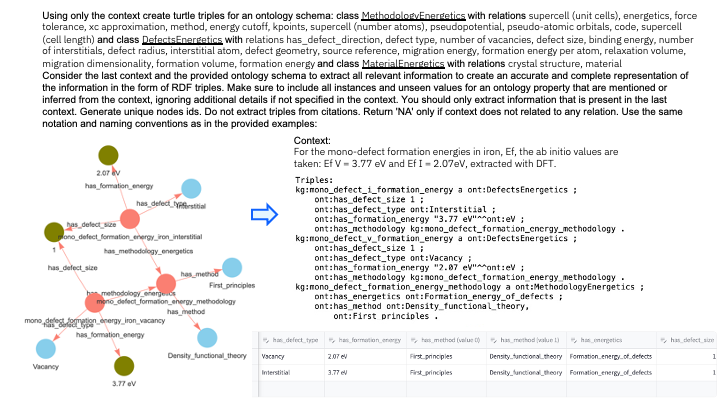}
\caption{Example of an automatically generated prompt with a detailed instruction, including an ontology-verbalization listing intermediate classes with their attribute and object relations, and a one-shot KG example, serialised in turtle, corresponding to the given passage (the Context) and compliant to the ontology. The corresponding KG for that passage is also visualized: intermediate nodes are shown in pink, leaf entities in blue and datatype literals in green. The sub-graph can also be presented as a table.}
\label{figprompt}
\end{figure}

In the ’simple’ version, the prompt just instructs the LLM to generate turtle triples for the ontology based solely on the context passage. The detailed instruction, placed after the ontology and before the examples, directs the model to consider accuracy and completeness, adhere to domain/range and functional constraints, ignore information not stated in the context, use consistent and unique URIs, avoid redundancy, and refrain from extracting triples from citations.
Providing the entire ontology graph requires a larger context window but explicitly informs the LLM of property constraints, functional relations (i.e., properties restricted to a single value), and short descriptions of datatype attributes and ontology classes. Including instances in addition to the schema can help the LLM resolve textual mention to correct URIs, essentially handling lexical matching to known instances.
An example of an ontology-verbalization prompt automatically generated from a configurable instruction, the ontology and one example is shown in Figure \ref{figprompt}. The same instruction is used when passing the full ontology, serialized in TTL, instead of the simple verbalization.

\begin{figure}[ht]
    \centering
    \includegraphics[width=\linewidth]{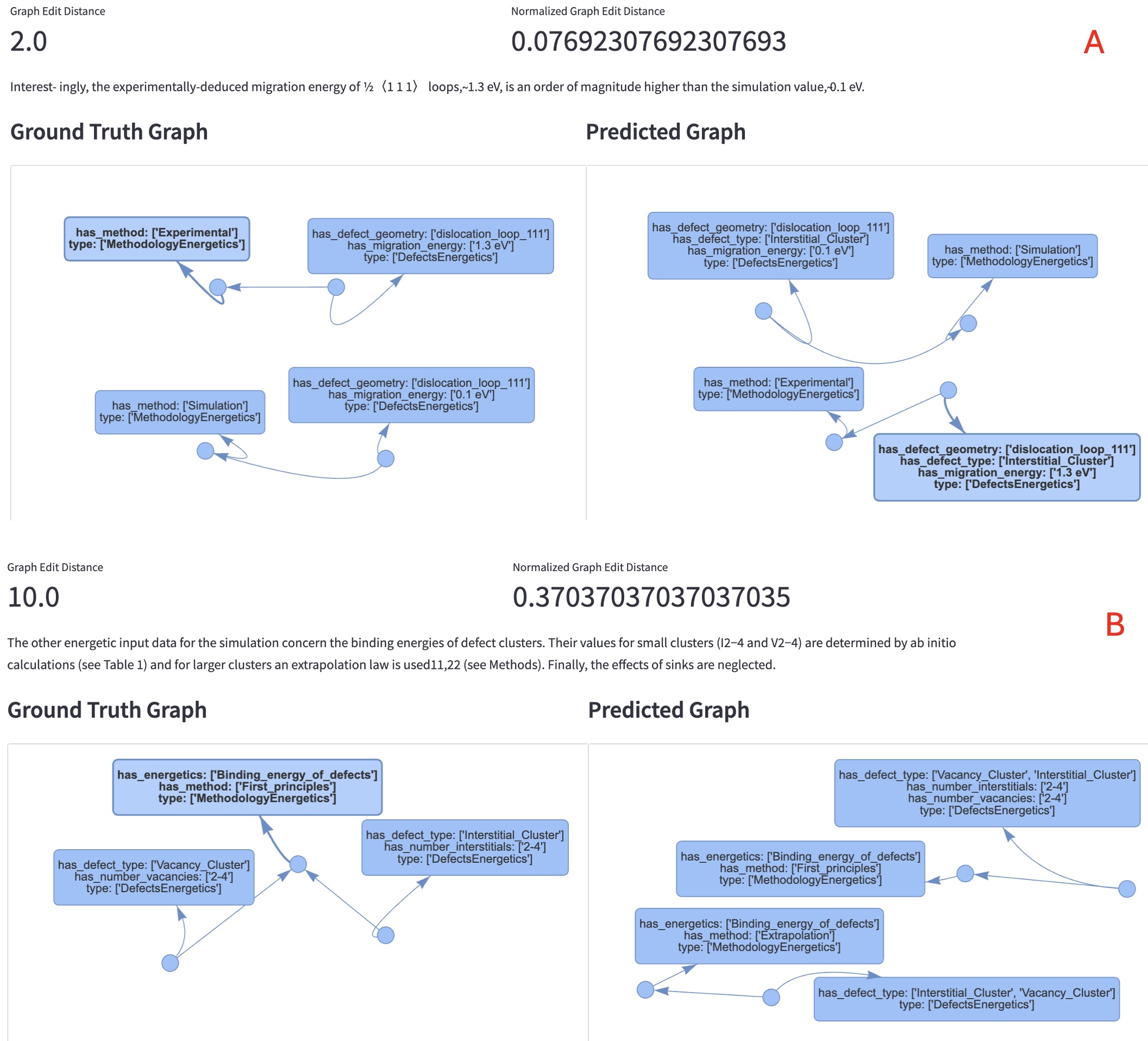}
    \caption{
    Predicted graphs generated by Mistral Large compared to ground truth examples.
    \textbf{(A)} Mistral-large zero-shot prediction using the ontology with instances require 4 edits (GED=4). Deviations include: using \textit{has\_defect\_size} instead of \textit{has\_number\_vacancies}, discrepancies in value formatting (e.g.: "$>4$" vs. "4–", "$>1$ eV" vs "1- eV"), and a missing \textit{has\_energetics} relation in the \textit{MethodologyEnergetics} node. These errors arise from the model not having seen examples illustrating conventions for representing ranges or for distinguishing when to use similar properties. 
    \textbf{(B)} Mistral Large in few-shot mode, even with access to the full ontology schema, fails to separate \textit{Interstitial\_Cluster} and \textit{Vacancy\_Cluster} into distinct \textit{DefectsEnergetics} nodes (GED = 10.0), violating the functional property constraint of \textit{has\_defect\_type}, which permits only one value per node. It also extracts extra nodes for large clusterswith a valid \textit{Extrapolation} method, which is correctly mentioned in the text but absent from the ontology and ground truth. 
}
    \label{fig:examples-2}
\end{figure}

Few-shot examples significantly enhance the semantic alignment and structural correctness of the extracted KGs with the ontology, especially for large passages with highly entangled multi-entity relationships. In zero-shot, errors arise because the LLM had not seen examples illustrating formatting conventions for representing values or how to distinguish between closely related properties. An example of this divergence for zero-shot settings can be seen in Figure \ref{fig:examples-2}.A.

Hallucinations - when the model fills in with plausible information not grounded in the input - is observed across all models, particularly in the zero-shot setting and for passages not providing enough data. In some cases, the model infers domain-reasonable triples that could be considered valid background (or axiomatic) knowledge, but are not explicitly mentioned in the text, and thus absent from the ground truth. Conversely, another common failure mode is the omission of relevant information explicitly present in the context — i.e., extracting partial information accurately but failing to capture the full set of relevant triples from the passage.

Detailed instructions help mitigate these issues by discouraging hallucination and requiring inclusion of all entities mentioned in context. In some cases, the models are able to infer more information from the sentence that it was omitted by our domain experts in the ground truth, but which is technically accurate and supported by the ontology schema. Examples can be seen in the methods extracted in Figure\ref{fig:examples-2}.B.
Detailed instructions guide the model to respect semantic constraints in the ontology, such as property domain and ranges and functional properties (i.e., those that can only have one value); while generally effective, models still occasionally generate KGs that violate ontology constraints, like in the example in Figure \ref{fig:examples-2}.B, where the functional constraint on \textit{has\_defect\_type} is not respected (a node should have been created for each defect type) 
Despite the ontology reducing ambiguity, variation in notation and phrasing persists (Figure~\ref{fig:examples-2}.A). Our evaluation metric penalizes deviations from the ground truth even when experts would consider the extracted values valid.

The few-shot samples obtained from input KG were manually selected to optimize coverage, ensuring that there is an example for nearly each property. We did not evaluate the effect of choosing different few-shot samples according to the model used; though prior work by Silva et al. \cite{dasilva2024automatedllmenabledextraction}, show that tailoring examples to each model can further enhance results. Moreover, as token limits become less of a concern, incorporating more examples is also likely to yield further gains \cite{JRC137500}. 

Overall, our results underscore the importance of prompt design in enabling LLMs extract structured knowledge reliably, guided by an ontology and few-shot examples to steer output quality in knowledge-intensive domains. Furthermore, the more capable LLMs could serve as teacher models to fine-tune smaller models, which can generate syntactically valid KGs but struggle with completeness and semantic-compliance. We leave the exploration of fine-tuning strategies for future work. 

\clearpage

\end{appendices}

\end{document}